\documentclass[preprint,12pt]{elsarticle}

\usepackage{amsmath,amssymb,amsfonts}
\usepackage{algorithmic}
\usepackage{graphicx}
\usepackage{algorithm,algorithmic}
\usepackage{hyperref}
\usepackage{textcomp}
\usepackage{adjustbox} 
\usepackage{bm} 
\usepackage{multirow} 
\usepackage{diagbox} 
\usepackage{comment} 
\usepackage{cleveref}
\usepackage{booktabs}
\usepackage{subcaption}
\usepackage{xcolor}

\usepackage{pifont} 
\newcommand{\cmark}{\ding{51}}%
\newcommand{\xmark}{\ding{55}}%

\usepackage[normalem]{ulem} 
\useunder{\uline}{\ul}{}

\journal{}

\begin{document}

\begin{frontmatter}

\title{Beyond a Single Mode: GAN Ensembles for Diverse Medical Data Generation}

\author[aff1]{Lorenzo Tronchin}

\author[aff2]{Tommy Löfstedt}

\author[aff1,aff3]{Paolo Soda}\corref{cor}
\ead{p.soda@unicampus.it, paolo.soda@umu.se}

\author[aff1]{Valerio Guarrasi}

\affiliation[aff1]{organization={Unit of Computer Systems and Bioinformatics, Department of Engineering \\ Università Campus Bio-Medico di Roma},
            city={Rome},
            country={Italy}}

\affiliation[aff2]{organization={Department of Computing Science \\ Umeå University},
            city={Umeå},
            country={Sweden}}

\affiliation[aff3]{organization={Department of Diagnostics and Intervention, Radiation Physics, Biomedical Engineering \\ Umeå University},
            city={Umeå},
            country={Sweden}}

\begin{abstract}
The advancement of generative AI, particularly in medical imaging, confronts the trilemma of ensuring high fidelity, diversity, and efficiency in synthetic data generation. While Generative Adversarial Networks (GANs) have shown promise across various applications, they still face challenges like mode collapse and insufficient coverage of real data distributions.  This work explores the use of GAN ensembles to overcome these limitations, specifically in the context of medical imaging. By solving a multi-objective optimisation problem that balances fidelity and diversity, we propose a method for selecting an optimal ensemble of GANs tailored for medical data. The selected ensemble is capable of generating diverse synthetic medical images that are representative of true data distributions and computationally efficient. Each model in the ensemble brings a unique contribution, ensuring minimal redundancy. We conducted a comprehensive evaluation using three distinct medical datasets, testing 22 different GAN architectures with various loss functions and regularisation techniques. By sampling models at different training epochs, we crafted 110 unique configurations. The results highlight the capability of GAN ensembles to enhance the quality and utility of synthetic medical images, thereby improving the efficacy of downstream tasks such as diagnostic modelling.
\end{abstract}


\end{frontmatter}

\section{Introduction}
\label{sec:introduction}
Generative Adversarial Networks (GANs)~\cite{goodfellow2014generative} have proven effective in various medical imaging tasks~\cite{chen2022generative,kebaili2023deep}.
However, synthetic data from GANs may fail due to mode collapse, memorisation of training data, and poor coverage of the real manifold~\cite{van2023beyond,xiao2021tackling}. 
The \textit{generative learning trilemma}~\cite{xiao2021tackling} includes three crucial requirements for practical applications: high-quality sampling (fidelity), mode coverage (diversity), and fast, computationally inexpensive sampling. 
Fidelity measures how well-generated samples resemble real ones~\cite{naeem2020reliable}, while diversity assesses the coverage of the variability in real samples~\cite{naeem2020reliable}.
GANs generate high-quality samples rapidly but have poor mode coverage~\cite{zhong2019rethinking}, often missing rare diseases or anomalies in medical imaging.

Precision and variability in synthetic data are critical in medical imaging, where diverse pathologies and patient-specific variations must be represented. This is essential for basic research, clinical training, and algorithmic validation, where realistic, diverse datasets develop robust diagnostic tools. Therefore, an ensemble of GANs can generate high-quality, diverse medical images, enhancing machine learning models' ability to generalise across unseen conditions~\cite{bib:caruso2024not,bib:caruso2024deep}.
A large body of literature~\cite{tolstikhin2017adagan,grover2018boosted,ghosh2018multi,hoang2018mgan,nguyen2017dual,mordido2018dropout,eilertsen2021ensembles,van2023synthetic,huang2017snapshot,wang2016ensembles} has investigated GAN ensembles to solve diversity issues. Approaches include boosting strategies~\cite{tolstikhin2017adagan,grover2018boosted}, multiple generators~\cite{ghosh2018multi,hoang2018mgan}, discriminators~\cite{nguyen2017dual}, or dropout~\cite{mordido2018dropout}. Eilertsen~\textit{et al.}~\cite{eilertsen2021ensembles} and Van Breugel~\textit{et al.}~\cite{van2023synthetic} evaluated ensemble performance on various datasets, showing improved results. However, they focused on two architectures, DCGAN~\cite{radford2015unsupervised} and PG-GAN~\cite{karras2017progressive}, without combining different architectures or sampling at various training steps. Huang~\textit{et al.}~\cite{huang2017snapshot} and Wang\textit{et al.}~\cite{wang2016ensembles} saved model parameters at different local minima but did not explore different architectures.
These approaches typically focus solely on generative model performance, neglecting the impact of synthetic data on downstream tasks. Our approach leverages pre-existing GANs, integrating multiple architectures and sampling at different training iterations to create a robust ensemble capturing real data complexity.
This work explores building a deep ensemble of GANs to recover as many training data modes as possible. We investigated GANs focusing on: (i) training different GANs (e.g., architecture, adversarial loss, regularisation) and (ii) sampling GANs at different epochs within a single model's training. We hypothesised that the min-max game dynamics~\cite{goodfellow2014generative} continuously change the generator network, allowing an ensemble of GANs based on the same initialisation but sampled at different epochs.
Our method is agnostic to the GANs included in the search space, helping answer the questions: \textit{``Which GAN should we use?''} and \textit{``When should we stop adversarial training?''}. We considered 22 GANs, covering a wide range of architectures~\cite{gulrajani2017improved,zhang2019self,brock2018large,karras2020analyzing}, conditioning methods~\cite{dumoulin2016learned,karras2019style,odena2017conditional,hou2022conditional,gong2019twin,miyato2018cgans,kang2021rebooting,lim2017geometric,kang2020contragan}, adversarial losses and regularisation~\cite{zhang2019self,mescheder2018training,chen2016infogan}, and differentiable augmentation modules~\cite{karras2020training,zhao2020differentiable}. We aimed to select a combination of GANs that maximised quality and minimised overlap, preventing mode collapse and redundancy.

This is the first extensive study on constructing GAN ensembles in the medical domain. Our ensemble selection method balances high-fidelity sample generation and spans the variability of real data with the smallest number of GANs necessary to ensure unique contributions. Our main contributions are:
\begin{enumerate}
    \item Introducing a new method to build GAN ensembles agnostic to the GAN model and training iterations, balancing data representation and diversity while minimizing overlap and redundancy.
    \item Systematically analysing the impact of different backbone architectures for embedding extraction in GAN evaluation, including variations of ImageNet-pre-trained models, fine-tuned to the medical domain, and testing unsupervised backbones.
    \item Reporting and analyzing extensive tests across three distinct medical datasets, considering the effect of the proposed ensemble method on downstream tasks and from fidelity and diversity perspectives. 
\end{enumerate}

The code for our experiments is available at: \url{https://github.com/ltronchin/GAN-Ensembles}.

\section{Methods}
\label{sec:methods}

\begin{figure*}[tb]
  \centering
   \includegraphics[width=\textwidth]{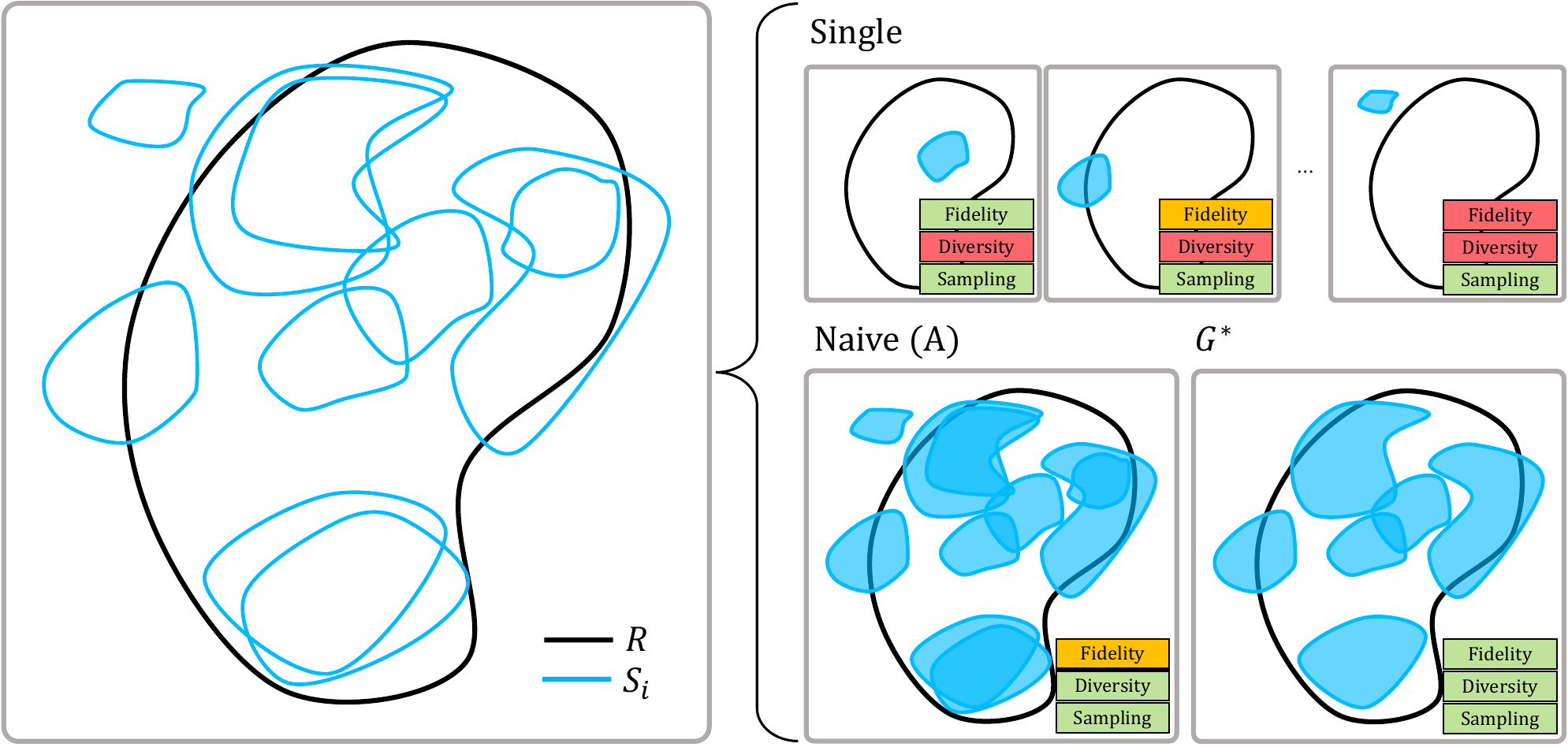}
   \caption{
   An illustration of the manifold occupied by real data, $R$, alongside the manifold covered by various GAN configurations, $S_i$.
\textit{Single}: Showcases three scenarios: $S_i$ entirely within the $R$, indicating high fidelity; $S_i$ partially overlapping with the $R$ representing medium fidelity; $S_i$ entirely outside the $R$ suggesting low fidelity.
Single GAN selection inherently has a low diversity due to limited real data coverage.
\textit{Naive (A)}: Depicts the collective space covered by all GANs, encompassing areas both inside and outside the $R$.
This approach results in medium fidelity and high diversity.
\textit{$G^*$}: Demonstrates the space covered by an ensemble of GANs selected through the proposed optimisation method, which aims to maximise coverage of the $R$ to ensure diversity discharging GAN outside the $R$, to ensure fidelity, using as few GANs as possible.}
   \label{fig:intution}
\end{figure*}

To select an appropriate GAN for data generation is a central decision that substantially impacts the quality and utility of the generated data, especially in the field of medical imaging where accuracy and variety in the data are paramount~\cite{bib:rofena2024deep}. A straightforward approach would be to select a single GAN from a pool of available models. 
However, this approach, while simple, often leads to sub-optimal results due to inherent trade-offs in fidelity and diversity inherent in any single GAN model.
A single GAN, depending on its architecture and training, may exhibit varying degrees of alignment with the real data space.
Some may generate data with high fidelity, others might partially overlap with the real data space offering a balance, while some might completely miss the mark, generating data with low fidelity. 
Moreover, relying on a single GAN inherently limits the diversity of the generated data, which is particularly detrimental in medical applications where the diversity of pathological features across different conditions must be well-represented~\cite{bib:guarrasi2023multi, bib:caruso2022multimodal, bib:guarrasi2022optimized, bib:guarrasi2022pareto, bib:guarrasi2021multi}.

An alternative, yet equally naive, would be to utilise all available GANs. 
While this method could theoretically maximise diversity, it may include models generating data outside the desired real data space, leading to a reduction of fidelity— a critical aspect when generating medical images where precision is crucial, incurring to high computational costs.
Recognising these limitations, we propose a method that selects an optimised ensemble.
By strategically selecting a subset of available GANs, the proposed method aims to harness the strengths of individual models while mitigating their weaknesses. 
The underlying intuition is that an ensemble can be tailored to maximise coverage of the real data space and minimise overlap among the GANs.
This approach strikes a balance between the extremes of single-model selection and the all-in approach, achieving an optimal trade-off between fidelity and diversity in the context of medical imaging.
The intuition behind the proposed method is illustrated in~\cref{fig:intution}.

In the proposed method, we aimed to harness the power of multiple GANs to create an optimal ensemble for generating medical imaging data.
Let $G = \{ G_1, G_2, \ldots, G_{|G|} \}$ denote the set of $|G|$ trained GANs, each generating sets of corresponding synthetic medical images $S_1, S_2, \ldots, S_{|G|}$, respectively, where $S_i:=S(G_i)$ are synthetic samples generated using the $i$th GAN.
We quantify the fidelity and diversity of these samples using a distribution quality metric $d$.
Fidelity is measured by an Intra-$d$ metric, which compares the generated samples of a single GAN and the real medical data.
Diversity is measured by an Inter-$d$ score comparing samples generated from different GANs.
For distribution fidelity metric, $d$, to compute Intra-$d$ and Inter-$d$, we used a combination of density (dns) and coverage (cvg)~\cite{naeem2020reliable} metrics, namely $d = 2\cdot\frac{\text{dns} \cdot \text{cvg}}{\text{dns} + \text{cvg}}$.
Density ensures that the generated samples densely populate the real data manifold, focusing on the core of the distribution rather than outliers (fidelity).
Coverage ensures the ensemble captures less frequent yet critical data features, such as rare anatomies in medical imaging (diversity).
To compute these metrics effectively, we leveraged Swapping Assignments between Views (SwAV)~\cite{caron2020unsupervised} as a backbone to extract the embeddings from the real training set and $|R_{Tr}|$ synthetic images.
We report further details on the distance metric, $d$, and the backbone in~\cref{app:Backbones_analysis_details}.

The objective is to find an ensemble $G^* \subseteq G$ of GANs that simultaneously maximises Intra-$d$ and minimises Inter-$d$.
Maximising Intra-$d$ ensures that the generated data is representative of real-world medical data by maintaining the quality and reliability of the synthetic images.
Conversely, Inter-$d$ ensures that each GAN in the ensemble contributes to diverse medical imaging data by avoiding redundancy in the data generated by different GANs and constraining the number of GANs in the resulting ensemble.
In fact, when training multiple independent GANs, some may generate similar synthetic data; therefore, we aim to select high-quality data while eliminating duplicates.

\begin{figure*}[tb]
    \centering
    \includegraphics[width=\textwidth]{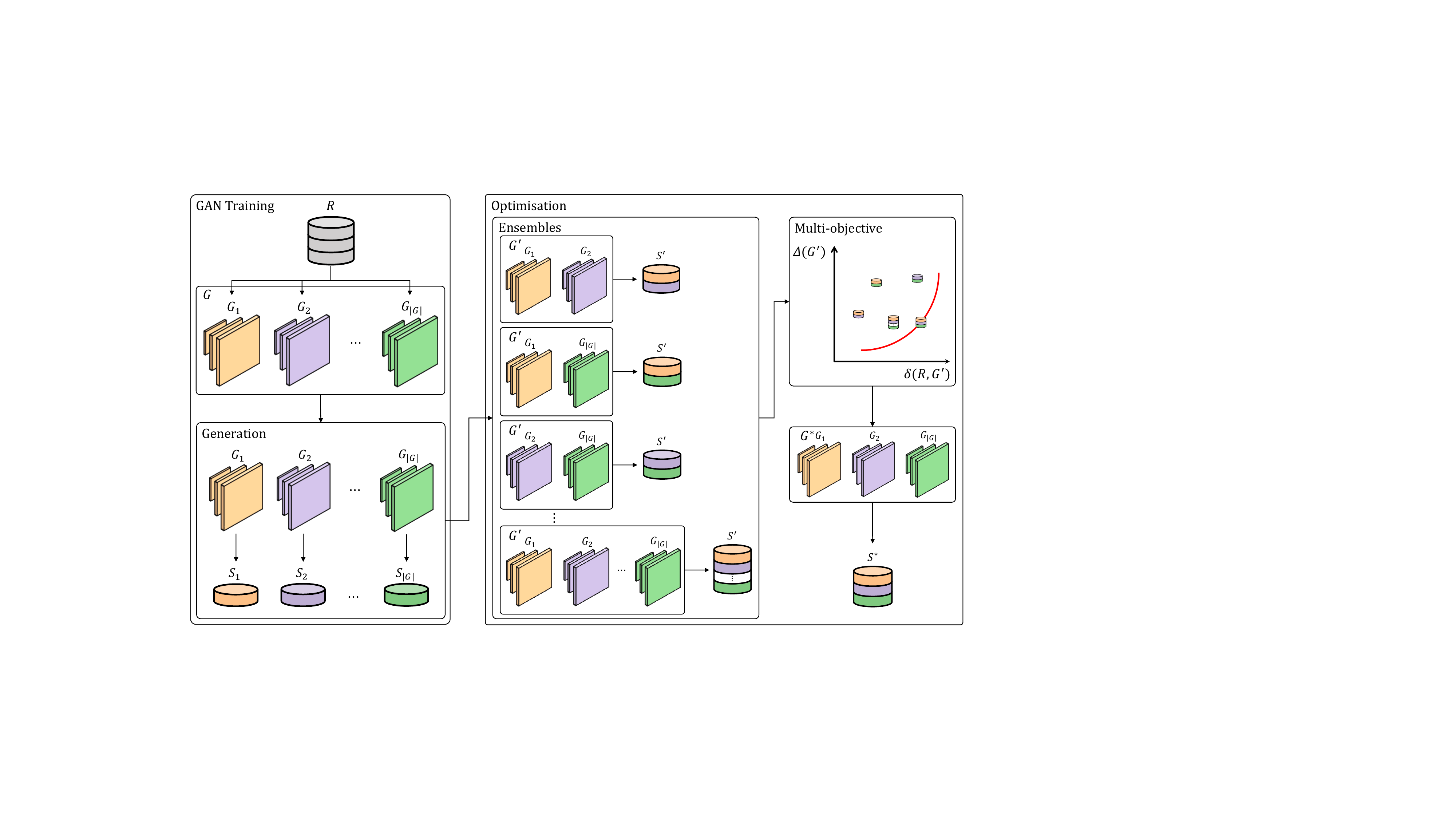}
    \caption{
This figure presents a step-by-step visualisation of the proposed methodology, comprising four main stages:
\textit{Training  GANs}: Multiple GAN architectures $G_i \in G$ are trained on a real data, $R$.
\textit{Generation}: Each GAN, $G_i$, is used to generate a synthetic dataset, $S_i$. These data are evaluated with respect to their fidelity and diversity.
\textit{Optimisation}: Various ensembles, $G'$, of GANs, are formed from the pool of trained models, followed by a multi-objective optimisation approach which maximises the closeness to the real data using $\delta$, and minimises the overlaps using $\Delta$. This results in a single \textit{best} ensemble, $G^*$, that is selected from the Pareto front and generates $S^*$.
}
    \label{fig:method}
\end{figure*}

\cref{fig:method} contains a schematic representation of the proposed method, which is composed of different steps described in the subsequent subsections.

\subsection{GAN Training}

Given a training set of real samples, $R$, multiple GANs $G_i \in G$, are trained to generate corresponding synthetic samples, $S_i := S(G_i)$, similar to those from the real data distribution.
Once all GANs are trained, we use the fidelity metric $d$, to evaluate a generic ensemble $G' \subseteq G$. 
The metric is used to quantify the difference between two data distributions.
As mentioned, we compute an Intra-$d$ (comparing real and synthetic distributions) and an Inter-$d$  metric (comparing synthetic distributions), now formally introduced.

Let $ x = [x_1, x_2, \ldots, x_{|G|}]$ be a vector that contains a binary decision variable for each GAN in an ensemble, such that $G' = G(x)=\{G_i \,|\, G_i \in G, x_i=1, \forall i=1,\ldots,|G|\}$. Hence, each $x_i$ indicates the presence of the corresponding $G_i$ in the ensemble $G$.

The Intra-$d$ is defined as the quality metric between a set of generated samples in an ensemble, $\{S(G_i) \,|\, G_i \in G'\}$, and the real data, $R$, as
\begin{equation}
    \delta(R, G') = d\bigg(R, \bigcup_{G_i \in G'} S(G_i)\bigg).
\end{equation}
The Intra-$d$ for a single GAN, $G_i$, is thus simply,
\begin{equation}
    \delta(R, \{G_i\}) = d\big(R, S(G_i)\big).
\end{equation}

The Inter-$d$ is defined as the average quality metric between all pairs of synthetic data generated by GANs in an ensemble, $G'$, as
\begin{equation}
    \Delta(G') = \frac{1}{|G'|(|G'|-1)} \sum_{\substack{G_i,G_j \in G' \\ G_i\neq G_j}} d\big(S(G_i), S(G_j)\big).
\end{equation}
The Inter-$d$ between a single pair of GANs, $G_i$ and $G_j$ is thus,
\begin{equation}
    \Delta\big(\{G_i, G_j\}\big) = d\big(S(G_i), S(G_j)\big).
\end{equation}

\subsection{Optimisation}

The aim is to find an ensemble, $ G^* \subseteq G $, that simultaneously maximises the Intra-$d$ and minimises the Inter-$d$.
The multi-objective (Pareto) optimisation problem that we want to solve is,
\begin{align*}
    \underset{x}{\text{minimise}} &\;
    \Big(%
        -\delta\big(R, G(x)\big),
        \Delta\big(G(x)\big)
    \Big) \\
    \text{subject to} &\; x_i \in \{0,1\}, \quad \forall i \in \{1,2,\ldots,|G|\}. \label{eq:optimisation}
\end{align*}
We thus seek a Pareto frontier in the trade-off between the two terms in the objective.
Through the optimisation process, we obtain the Pareto front, $P$, a set of Pareto optimal solutions (a set of ensembles), each characterised by a unique trade-off between the Intra-$d$ and Inter-$d$ objectives.
All solutions, $x\in P$, are Pareto optimal since there does not exist another solution, $x'\notin P$, such that~\cite{bib:deb2002fast},
\begin{equation}
    \delta\big(G(x')\big) \leq \delta\big(R, G(x)\big)
    \quad \text{and} \quad
    \Delta\big(G(x')\big) \geq \Delta\big(G(x)\big),\label{eq:pareto_condition}
\end{equation}
for any $x\in P$, with at least one inequality being strict.

We select a final \textit{best} ensemble of GANs, $G^*\in P$, as
\begin{equation}
    G^* = \underset{x \in P}{\arg\max} \; \delta\big(R, G(x)\big).
    \label{eq:selection_criteria}
\end{equation} 
In this way, we prioritise solutions that closely approximate the real data distribution, being vital for the utility of the data generation in downstream tasks.

From the final \textit{best} Pareto optimal ensemble, $G^*$, we generate a synthetic dataset,
\begin{equation}
    S^* := \bigcup_{G_i \in G^*} S(G_i),
\end{equation} \label{eq:optimal_pareto}
which aggregates the data generated from all GANs $G_i \in G^* $.

\section{Experimental Setup}
\label{sec:exp_setup}

We use three medical imaging datasets with different characteristics: PneumoniaMNIST~\cite{yang2023medmnist}, BreastMNIST~\cite{yang2023medmnist}, and AIforCOVID~\cite{soda2021aiforcovid}.
\cref{tab:datasets} summarises the main characteristics of each dataset. 
We trained 22 GANs for 100\,000 iterations for each dataset. 
We uniformly sampled each model every 20\,000 training iteration, i.e., 5 samples, resulting in a search space of 110 models.
Note that we trained the GANs only on the training set from each dataset to avoid bias in the generation process.
We report further details on the datasets and the GANs in the following subsections.

\begin{table}[tb]
\caption{Summary of the medical datasets. \bm{$|R_{Tr}|$}, \bm{$|R_{Vl}|$} and \bm{$|R_{Te}|$} denote the number of samples in the training, validation and test set, respectively.}
\label{tab:datasets}
\centering
\begin{adjustbox}{width=\columnwidth}
\begin{tabular}{lccccccc}
\toprule
 {\textbf{Dataset}} & {\textbf{Modality}} & \bm{$|R_{Tr}|$} & \bm{$|R_{Vl}|$} & \bm{$|R_{Te}|$} & {\textbf{Task}} \\
\midrule
PneumoniaMNIST~\cite{kermany2018identifying} & X-ray  & 4\,708 &  524  & 624 & Normal, Pneumonia \\
BreastMNIST~\cite{al2020dataset} & Ultrasound    & \phantom{1}546         & \phantom{1}78          & 156         & Normal or Benign, Malignant \\
AIforCOVID~\cite{soda2021aiforcovid}& X-ray & \phantom{1}664  & \phantom{1}74 & \phantom{1}82  & Mild, Severe \\
\bottomrule
\end{tabular}%
\end{adjustbox}
\end{table}

\subsection{Datasets}
\label{app:Datasets}
We used three medical imaging datasets with different characteristics: PneumoniaMNIST~\cite{yang2023medmnist}, BreastMNIST~\cite{yang2023medmnist}, and AIforCOVID~\cite{soda2021aiforcovid}.

PneumoniaMNIST includes 5856 pediatric chest X-ray images for the binary classification of pneumonia versus normal cases~\cite{kermany2018identifying}.
We divide the original training set in a 9:1 ratio for our training and validation while utilising the original validation set as our test set.
The source images are grayscale and of various sizes. 
We center-cropped the images with a window size of the length of the short edge and resized them to $224 \times 224$.

BreastMNIST consists of 780 breast ultrasound images from the original dataset that and categorised into three classes: normal, benign, and malignant~\cite{al2020dataset}. 
We simplified this into a binary classification task by combining normal and benign as one class and malignant as the other.
We split the data into a 7:1:2 ratio for training, validation, and test sets.
We resized the images from $ 500 \times 500$ to $224 \times 224$.
For the preprocessing steps for PneumoniaMNIST and BreastMNIST, we followed the guidelines from Yang et al.~\cite{yang2023medmnist}.

AIforCOVID~\cite{soda2021aiforcovid} includes X-ray scans of 820 patients recorded from six different Italian hospitals.
The patients' data were collected at the time of hospitalization if the TR-PCR test resulted positive to the SARS-CoV-2 infection. 
The dataset categorises patients into two distinct groups based on the severity of their treatment outcomes, i.e. mild or severe.
The mild class includes patients who
were either in domiciliary isolation or hospitalised without ventilatory support; while the severe category includes patients who required non-invasive ventilation,
the admission to intensive care, or deceased.
We split the data into an 8:1:1 ratio for training, validation, and test sets.
We processed the X-ray by extracting the segmentation mask of lungs, using a U-Net~\cite{ronneberger2015u} trained on two non-COVID-19 lung datasets~\cite{shiraishi2000development,jaeger2014two}. 
Then we used the mask to extrapolate the minimum squared bounding box containing both lungs. The extracted box was then resized to $224 \times 224$.

\subsection{Rationale behind the choice of GANs}
\label{app:rationale_gans}

We tested a total of 22 different GANs, divided into the six categories presented in~\cref{tab:selected_gans}. 
From left to right columns, we differentiated according to the model architecture, conditioning goal for the generator and discriminator, adversarial loss, regularisation, and data-efficient training.

\begin{table}[!ht]
\centering
\begin{adjustbox}{width=\columnwidth}
\begin{tabular}{lcccccc}
\toprule
\textbf{Method} &
  \textbf{Architecture} &
  \textbf{\begin{tabular}[c]{@{}c@{}}Conditioning goal \\ generator\end{tabular}} &
  \textbf{\begin{tabular}[c]{@{}c@{}}Conditioning goal \\ discriminator\end{tabular}} &
  \textbf{Adversarial loss} &
  \textbf{Regularisation} &
  \textbf{Data-efficient} \\
  \midrule
ACGAN-Mod         & ResNetGAN        & cBN    & AC    & Hinge    & ---            & ---         \\
ACGAN-Mod-ADC     & ResNetGAN        & cBN    & ADC   & Hinge    & ---            & ---         \\
ACGAN-Mod-TAC     & ResNetGAN        & cBN    & TAC   & Hinge    & ---            & ---         \\
SNGAN             & ResNetGAN        & cBN    & PD    & Hinge    & SNd            & ---         \\
SAGAN             & ResNetGAN-Att    & cBN    & PD    & Hinge    & SNd, SNg       & ---         \\
BigGAN            & ResNetBigGAN-Att & cBN    & PD    & Hinge    & SNd, SNg       & ---         \\
BigGAN-Info       & ResNetBigGAN-Att & cBN    & PD    & Hinge    & SNd, SNg, Info & ---         \\
BigGAN-ADA        & ResNetBigGAN-Att & cBN    & PD    & Logistic & SNd, SNg       & ADA         \\
BigGAN-DiffAug    & ResNetBigGAN-Att & cBN    & PD    & Hinge    & SNd, SNg       & DiffAugment \\
ReACGAN           & ResNetBigGAN-Att & cBN    & D2D-CE & Hinge    & SNd, SNg       & ---         \\
ReACGAN-Info      & ResNetBigGAN-Att & cBN    & D2D-CE & Hinge    & SNd, SNg, Info & ---         \\
ReACGAN-ADA       & ResNetBigGAN-Att & cBN    & D2D-CE & Logistic & SNd, SNg       & ADA         \\
RaACGAN-DiffAug   & ResNetBigGAN-Att & cBN    & D2D-CE & Hinge    & SNd, SNg       & DiffAugment \\
ReACGAN-ADC       & ResNetBigGAN-Att & cBN    & ADC   & Hinge    & SNd, SNg       & ---         \\
ReACGAN-TAC       & ResNetBigGAN-Att & cBN    & TAC   & Hinge    & SNd, SNg       & ---         \\
MHGAN             & ResNetBigGAN-Att & cBN    & MH    & MH       & SNd, SNg       & ---         \\
ContraGAN         & ResNetBigGAN-Att & cBN    & 2C    & Logistic & SNd, SNg       & ---         \\
StyleGAN2         & StyleGAN2        & cAdaIN & SPD   & Logistic & R1             & ---         \\
StyleGAN2-Info    & StyleGAN2        & cAdaIN & SPD   & Logistic & R1, Info       & ---         \\
StyleGAN2-D2D-CE   & StyleGAN2        & cAdaIN & D2D-CE & Logistic & R1             & ---         \\
StyleGAN2-ADA     & StyleGAN2        & cAdaIN & SPD   & Logistic & R1             & ADA         \\
StyleGAN2-DiffAug & StyleGAN2        & cAdaIN & SPD   & Logistic & R1             & DiffAugment \\
\bottomrule
\end{tabular}%
\end{adjustbox}
\caption{Implemented GANs. cBN: conditional Batch Normalization. cAdaIN: Conditional version of Adaptive Instance Normalization. AC: Auxiliary Classifier. PD: Projection Discriminator. TAC: Twin Auxiliary Classifier. SPD: Modified PD for StyleGAN. 2C: Conditional Contrastive loss. MH: Multi-Hinge loss. ADC: Auxiliary Discriminative Classifier. D2D-CE: Data-to-Data Cross-Entropy. R1: R1 regularisation, SNd and SNd: Spectral Normalization for the generator and discriminator, respectively. Info: Information-theoretic regularisation. ADA: Adaptive Discriminator Augmentation. DiffAugment: Differentiable Augmentation.}
\label{tab:selected_gans}
\end{table}

\subsubsection{Architecture}
We cover a wide range of GAN architectures proposed in the literature, i.e., ResNet-style generator and discriminator \cite{gulrajani2017improved}, adding the self-attention layer \cite{zhang2019self} and scaling up with the hierarchical embeddings \cite{brock2018large}. 
Moreover, we include StyleGAN2~\cite{karras2020analyzing}, which is known for generating high-quality images.
We report the architecture tested for each GAN in the second column of~\cref{tab:selected_gans}.

\subsubsection{Conditional image generation}
We aimed to use synthetic images generated by GANs for downstream classification tasks.
Thus, we only included in our study conditional GANs, i.e., models can generate images conditioned on class labels
We conditioned the generation according to the labels associated with each image in the real training set.
In the second and third columns of~\cref{tab:selected_gans}, we report the conditioning goals for the generator and discriminator, respectively.
To implement the conditional generator, we utilised conditional Batch Normalization \cite{dumoulin2016learned} and a conditional version of Adaptive Instance Normalization \cite{karras2019style} when training the StyleGAN-based architectures.
The former provides class information within the batch norm layer, while the latter exploits the Ada layer at each convolutional generator block.
To train the conditional generator, a lot of effort was put into effectively injecting the conditional information into the discriminator or auxiliary classifier that guides the conditional generator.
Auxiliary Classifier Generative Adversarial Network \cite{odena2017conditional} has been widely used due to its simplicity.
It utilises an Auxiliary Classifier (AC) that first attempts to recognise the labels of data and then teaches the generator to produce label-consistent data.
However, it has been reported that AC-GAN suffers from a low intra-class diversity problem in the generated samples~\cite{hou2022conditional}.
Thus, we tested different conditioning techniques that aim to cope with this problem, such as Auxiliary Discriminative Classifier \cite{hou2022conditional}, Twin Auxiliary Classifier \cite{gong2019twin}, Projected Discriminator \cite{miyato2018cgans} and its modified version for StyleGAN-based architectures (SPD),  Data-to-Data Cross-Entropy \cite{kang2021rebooting}, Multi-Hinge loss \cite{lim2017geometric} and Conditional Contrastive loss \cite{kang2020contragan}.
We report the different conditioning techniques in columns three and four of~\cref{tab:selected_gans}.

\subsubsection{Adversarial loss and regularisation}
To cope with the unstable nature of GAN training, many researchers have proposed different adversarial losses and regularisation techniques.
We refer to the adversarial loss and regularisation of SA-GAN~\cite{zhang2019self} as a baseline.
It uses the hinge loss (Hinge in~\cref{tab:selected_gans}) as the GAN objective and Spectral Normalization for the discriminator (SNd) and the generator (SNg). 
The StyleGAN-based architectures use logistic loss (Logistic~\cref{tab:selected_gans}) and R1-regularisation~\cite{mescheder2018training}.
Finally, for both streams, StyleGAN-based and non-StyleGAN-based, we test the effect of information-theoretic regularisation~\cite{chen2016infogan} (Info in~\cref{tab:selected_gans}.
A summary of the tested adversarial losses and regularisation techniques can be found in columns five and six of~\cref{tab:selected_gans}.

\subsubsection{Techniques for data-efficient Training}
The GAN discriminator is prone to memorising the training dataset and presenting the authenticity score of a given image without considering the realism of the created images~\cite{karras2020training}.
Thus, researchers propose to apply data augmentations on real and fake images to prevent the discriminators from overfitting.
We investigated two approaches: Differentiable Augmentation \cite{zhao2020differentiable} and Adaptive Discriminator Augmentation \cite{karras2020training} as shown in the last column of~\cref{tab:selected_gans}.

\subsubsection{Same GAN, different training iterations}
Given the inherently unstable nature of adversarial training of GANs, we hypothesised that sampling the same GAN model at different iterations of training can yield variations of the model, each capturing different modes of the real data.
This understanding is critical as it would suggest that variability in GAN outputs is not solely dependent on the model architecture but is also significantly influenced by the training iteration at which the model is sampled.
We, therefore, explored the diversity within a single GAN model across different stages of its training.
To this end, we trained each of the 22 different GAN models for 100\,000 iterations, sampling a model at every 20\,000 iterations.
This results in five samples of each GAN model.
This strategy does not require further training as these instances are sampled while training the same GAN model, i.e., we obtain five models to add in the ensemble for every GAN.
We define the search space for the Pareto ensemble search as the union of the 22 GANs with samples at the five different training steps, resulting in a total of 110 GANs.

\subsubsection{GAN Training}
For GAN training, we resized the training images according to the dimensions presented for each dataset in~\cref{app:Datasets}.
We employed the Lanczos filter for interpolation to minimise aliasing artefacts. 
We normalised the images to the range $[-1,1]$.
The images were saved to disk each 20\,000 steps for evaluation purposes, using a lossless compression format (TIFF), and without using any quantisation operations.
We refer to \url{https://github.com/ltronchin/GAN-Ensembles} for the detailed training setting for each GAN.

\section{Results and Discussions}

\subsection{Impact of $G^{*}$  on Downstream Tasks}
\label{sec:Impact_downstream}
We evaluated the effectiveness of the proposed ensemble method, designing different experiments that trained a ResNet18~\cite{he2016deep} from scratch.
First, to set a baseline, we trained the ResNet18 on the real training data, $R$ (Real).

Then, we considered the ResNet18 when training on the synthetic data, $S_i$, generated from each $G_i$.
This resulted in 110 separate training procedures of the downstream task due to a search space of 110 GANs models (different $G_i$ architectures, adversarial loss, training step).
These experiments provided insight into the model's performance when trained on synthetic data generated by individual GANs, offering a perspective on the variability and effectiveness of each GAN for the downstream task.
We ranked the GANs based on the performance of the ResNet18 model on the real test set.
Accordingly, we outlined the highest-performing GAN as Single (top-1), the average performance of the top five GANs as Single (top-5) and the average across all GANs as Single (average).
This analysis clarifies the performance differences when using synthetic data from different GANs, showing the utility of the synthetic data from each GAN for real-world applications.
We reported the real test set performances achieved by the ResNet18 when trained with the synthetic dataset from the single GANs in~\cref{tab:single_gan_XRay,tab:single_gan_ultrasound,tab:single_gan_aiforcovid}, in~\cref{app:single_gan_analysis},
highlighting the top-1 and top-5 GANs in green and yellow for each dataset.

We further designed two additional experiments that performed no selection on the available GANs. 
The former trained the ResNet18 on the synthetic data $S = \bigcup_{G_i\in G} S(G_i)$ generated by all the 110 $G_i$ (Naive (A)).
The latter exploited the synthetic data generated by a randomly selected subset of GANs $G' \subseteq G$ (Naive (R)).

Finally, we trained the ResNet18 using the data generated by the optimally selected ensemble $G^{*}$, i.e., using $S^{*}$. In~\cref{tab:single_gan_XRay,tab:single_gan_ultrasound,tab:single_gan_aiforcovid}, in~\cref{app:single_gan_analysis}, we highlighted with a star the GANs selected by our multi-optimisation procedure.
Additionally, selecting the best ensemble among the Pareto-optimal solutions, as defined in~\cref{eq:optimal_pareto}, may not always result in the optimal outcome.
To address this, we trained the ResNet18 model on all Pareto-optimal solutions. 
We then selected the solution that achieved the highest geometric mean (g-mean) on the real test set, referring to this as the Oracle.

We fixed the number of training images for all experiments to the dimensionality of the real training set $|R_{Tr}|$ to allow direct comparison (check~\cref{tab:datasets} for the size of each training dataset).
By fixing the length of the training dataset to $|R_{Tr}|$, our procedure does not add a time overhead for training the downstream model. 
Indeed, being the total number of synthetic images for training always the same, the number of images generated by each GAN is a fraction of the total synthetic set, i.e., $\frac{|R_{Tr}|}{\#G_i}$, where $\#G_i$ represents the number of GAN models in the ensemble, fixing the overall sampling time.
Rather, including multiple $G_i$ in the solution impacts the GPU memory footprint as more generators need to be saved to the device.
To consider the impact of using the ensemble on GPU, we highlighted the number of $G_i$ used for each experiment. 
For a direct comparison, we set $\#G_i$ to be the same in both Naive (R) and $G^{*}$.
We introduced a new metric, $\gamma_{RS}$, to evaluate the efficacy of synthetic data, $S'$, compared to real training data, $R$, when training the ResNet18.
We defined $\gamma_{RS}$ as the performance gap, in percentage, between training the ResNet18 with the real training data, $R$, and the synthetic data, $S'$, $\gamma_{RS} = \left( \frac{\text{g-mean}(S') - \text{g-mean}(R)}{\text{g-mean}(R)} \right) \cdot 100$.
Moreover, to account for the variability inherent in the training process, we repeatedly trained the ResNet18 20 times.
Thus, for the performance metrics computed on the real test set, we show the mean and standard deviation across the 20 trained ResNet18 models.
We chose g-mean as our performance metric to account for potential imbalances in the test set, a common scenario in medical datasets.
We provide more details about the ResNet18 hyperparameters in~\cref{app:training_evaluation_setup}.

\begin{table*}[tb]
\caption{Downstream model performances on the real test set. 
For each experiment we reported the mean g-mean and the standard error across the 20 training repetitions, $\gamma_{RS}$, i.e., the \% gap between training the downstream model with the real and synthetic data and the number of GANs used in the solution, $\#G_i$. 
The best results are highlighted in bold.
*Results obtained ranking GAN's according to the downstream task performances on the real test set.}
\label{tab:downstream_task}
\centering
\begin{adjustbox}{width=\textwidth}
\begin{tabular}{lccccccccc}
\toprule
          & \multicolumn{3}{c}{\textbf{PneumoniaMNIST}}                      & \multicolumn{3}{c}{\textbf{BreastMNIST}}              & \multicolumn{3}{c}{\textbf{AIforCOVID}}                   \\
 & g-mean $\uparrow$ & $\gamma_{RS}$ $\downarrow$ & $\#G_i$ $\downarrow$ & g-mean $\uparrow$ & $\gamma_{RS}$ $\downarrow$ & $\#G_i$ $\downarrow$ & g-mean $\uparrow$ & $\gamma_{RS}$ $\downarrow$ & $\#G_i$ $\downarrow$ \\
\midrule
Real      & $0.822 \pm 0.017$           & ---            & ---   & $ 0.817 \pm 0.024$  & ---                    & ---   & $ 0.607 \pm 0.036 $   &---                     & ---   \\
\hline
Single (top-1)* & $0.854 \pm 0.007 $    &   $ +3.9 $                & 1   & $ 0.707 \pm 0.028$     &  $ -13.5 $             & 1   & \boldmath{$ 0.588 \pm 0.032$}    &  \boldmath{$-3.1$}          & 1   \\
Single (top-5)* & $0.842 \pm 0.010 $    &   $ +2.4$                & 1   & $ 0.697\pm 0.026$    &  $-14.7 $             & 1   & $ 0.555 \pm 0.039 $    &  $   -8.6   $          & 1   \\
Single (average)  & $0.652 \pm 0.038$    &   $-20.7$                & 1   & $ 0.407 \pm 0.053 $     &  $-50.2 $             & 1   & $ 0.339 \pm 0.075 $    &  $ -44.2       $          & 1   \\
\hline
Naive (R) & $ 0.822 \pm 0.017 $   & $0.0      $           & 38  & $ 0.664 \pm 0.038 $    &     $-18.7  $          & 32  & $ 0.533 \pm 0.030 $            & $-12.2      $     & 30  \\
Naive (A) & $0.823 \pm 0.016$    &   $+0.1 $                & 110 & $ 0.714 \pm 0.038 $     &  $-12.6      $        & 110 & $ 0.487 \pm 0.063 $     &$ -19.8         $         & 110 \\
\hline
$G^{*}$ (our)   & $0.867 \pm 0.014 $  &     $  +5.5     $         & 38  & \boldmath{$0.755 \pm 0.041$} & \boldmath{$-7.6$} & 32  & $ 0.573 \pm 0.036  $ & $-5.6$ & 30  \\
$G^{*}$ (Oracle)    & \boldmath{$0.881 \pm 0.014$} &  \boldmath{$+7.2$} & 40  & \boldmath{$ 0.755 \pm 0.041 $}  &  \boldmath{$-7.6$}                 & 38  & $ 0.573 \pm 0.036$     &  $-5.6$              & 30  \\
\bottomrule
\end{tabular}%
\end{adjustbox}
\end{table*}

\cref{tab:downstream_task} shows the results of the experiments, where we reported in rows two to four the results for Single experiments.
Notably, for PneumoniaMNIST, both Single (top-1) and Single (top-5) configurations yield a positive $\gamma_{RS}$ values, indicating improved performance. 
Conversely, we attained a negative $\gamma_{RS}$ for BreastMNIST and AIforCOVID datasets.
The Single (top-1) experiment is determined by ranking the GANs according to ResNet18 real test set g-mean. 
Thus, it underscores the Oracle solution when training the downstream model with the synthetic data from a single GAN.
We found StyleGAN-ADA to be the best architecture for all the datasets (green in~\cref{tab:single_gan_XRay,tab:single_gan_ultrasound,tab:single_gan_aiforcovid}).
Moreover, considering the top-5 architecture, i.e., GAN's whose synthetic data achieved the Single (top-5) test set g-mean, we discovered the best solution to be skewed towards style-based architectures (yellow in~\cref{tab:single_gan_XRay,tab:single_gan_ultrasound,tab:single_gan_aiforcovid}).
These findings confirm the StyleGAN-based models as the gold standard for synthetic data generation in low-data regimes~\cite{karras2020training}.
We found no clear evidence about the best iteration to stop the adversarial training, confirming that the challenge of properly setting the GAN's hyper-parameters is still an open issue to be addressed.
For the interested reader, we deepen the single GAN analysis in~\cref{app:single_gan_analysis}.
Comparing the Single (top-1) and Single (average) experiments in the second and fourth rows, we observed a decrease in $\gamma_{RS}$ values: from $+3.9\%$, $-13.5\%$, and $-3.1\%$ to $-20.7\%$, $-50.2\%$, and $-44.2\%$, for PneumoniaMNIST, BreastMNIST, and AIforCOVID, respectively.
This is due to using $S_i$ for training the ResNet18 may suffer numerous failures that can occur in adversarial training, like mode collapse, overfitting, or, more straightforwardly, limited diversity/mode coverage. 
This underscores the complexity of selecting a GAN to maximise the utility of synthetic images in downstream tasks.

Comparing the last and second-to-last rows ($G^{*}$ and Oracle experiments) against the rows two to four (Single experiments) in~\cref{tab:downstream_task}, we found that, overall, $G^{*}$ outperforms the single GAN solution, showing its capability to recover real data modes effectively, i.e., increasing the diversity.
Indeed, even if $G^{*}$ does not always beat the real data training performances, the real-synthetic gap, $\gamma_{RS}$, is reduced.
In particular, we have a $\gamma_{RS}$ of $-7.6\%$ and $-5.6\%$ to $-50.2\%$ and $-44.2\%$ when comparing the $G^{*}$ and the Single (average) experiment performances for the BreastMNIST and AIforCOVID datasets. 
For the PneumoniaMNIST dataset, our method further increased the real test set g-mean to the Single (top-1) experiment with a positive real-synthetic gap of $+5.5\%$.
Considering the second-to-last and second rows, $G^{*}$ and Single (top-1) experiments in~\cref{tab:downstream_task}, our method outperforms the StyleGAN2-ADA g-mean for PneumoniaMNIST and BreastMNIST while obtaining worse g-mean for AIforCOVID.
Moreover, we noticed that the optimal solution chosen by~\cref{eq:selection_criteria} ($G^{*}$), from among the Pareto front solutions in $P$, aligns with the Oracle for two out of three datasets.
This finding validates the approach to choose the solution that maximises the  Intra-$d$ term, according to~\cref{eq:selection_criteria}, i.e., prioritise solutions that closely approximate the real data distribution.
We report a deeper analysis of the Pareto frontier in~\cref{app:Backbones_analysis_pareto}, where we display in~\cref{fig:pareto_plots_swav} the Pareto plots for the PneumoniaMNIST, BreastMNIST, and AIforCOVID datasets.

Turning now the attention to rows five and six in~\cref{tab:downstream_task}, i.e., Naive (R) and Naive (A) experiments, respectively, we found that assembling different $G_i$ increased the downstream task performances on the synthetic datasets with respect to the average scores achieved by the single GANs (Single (average)) in all cases, but achieved the worst performances concerning our solutions.
We achieved a $\gamma_{RS}$ of $0.0\%$, $-18.7\%$ and $-12.2\%$ for the Naive (R) approach, while a $\gamma_{RS}$ of $+0.1\%$, $-12.6\%$ and $-19.8\%$  for the Naive (A) approach, for PneumoniaMNIST, BreastMNIST and AIforCOVID, respectively.
These results show the dependence of the Naive (R) and Naive (A) approaches on the performances of single $G_i$s in the search space.
Indeed, randomly sampling from the 110 GANs in the search space (Naive (R)) or naively selecting all the available $G_i$s (Naive (A)) yields an increased probability of ensembling low fidelity and/or diversity models, reducing the ensemble utility in the downstream task.
Moreover, the Naive (A) approach has a GPU footprint three times higher with respect to our solution, as emerges considering the number of GANs in the ensemble $\#G_i$ for the Naive (A) and the $G^*$ rows in~\cref{tab:downstream_task}.
Specifically, the Naive (A) employs 110 GANs across all datasets, while $G^*$ uses 38, 32 and 30 GANs for PneumoniaMNIST, BreastMNIST, and AIforCOVID, respectively.

To test the generalisability of our method across different search spaces, we limited the number of different GAN models employed.
We varied the number of GAN models from 3 to 7 with a step of 1, and for each experiment, we randomly sampled from the initial 22 GAN model architectures pool.
We uniformly sampled every 20 000 training iterations, resulting in 15, 20, 25, 30 and 35 models.
We used our multi-objective optimisation to identify the Pareto optimal solution for each configuration and trained the ResNet18 model on the synthetic dataset.
\begin{table*}[tb]
\caption{Downstream model performances on the real test set. For each experiment, we reported the mean g-mean and the standard error across the 20 training repetitions and the number of GANs used in the solution, $\#G_i$. 
We denote with $|G|_{M}$, $|G|_{I}$ and $|G|$ the number of different models, training steps per model, and total GANs in the search space, respectively.
The best results are highlighted in bold.}
\label{tab:limited_search_space}
\centering
\begin{adjustbox}{width=\textwidth}
\begin{tabular}{lccccccccc}
\toprule
      & & & & \multicolumn{2}{c}{\textbf{PneumoniaMNIST}} & \multicolumn{2}{c}{\textbf{BreastMNIST}} & \multicolumn{2}{c}{\textbf{AIforCOVID}} \\
      &$|G|_{M}$  & $|G|_{I}$  &  $|G|$ & g-mean  $\uparrow$             & $\#G_i$ $\downarrow$    & g-mean     $\uparrow$       & $\#G_i$ $\downarrow$   & g-mean $\uparrow$          & $\#G_i$ $\downarrow$   \\
      \midrule
Real  & ---  & --- &  --- &  $0.822 \pm 0.017$ & --- & $0.817 \pm 0.024$ & --- & $0.607 \pm 0.036$ & --- \\
\midrule
\multirow{5}{*}{$G^{*}$ (our)} & 3 & \multirow{5}{*}{5} & 15 & $0.830 \pm 0.018$ & 4 & $0.655 \pm 0.039$ & 3 & $0.540 \pm 0.037$ & 3 \\
& 4 & & 20 & $0.851 \pm 0.017$ & 5 & $0.730 \pm 0.038$ & 4 & $0.498 \pm 0.036$ & 4 \\
& 5 & & 25 & $0.856 \pm 0.014$ & 7 & $0.724 \pm 0.037$ & 6 & $0.558 \pm 0.035$ & 5 \\
& 6 & & 30 & \boldmath{$0.860 \pm 0.015$} & 9 & $0.735 \pm 0.035$ & 7 & $0.570 \pm 0.034$ & 6 \\
& 7 & & 35 & $0.853 \pm 0.016$ & 9 & \boldmath{$0.737 \pm 0.036$} & 9 & \boldmath{$0.570 \pm 0.033$} & 6 \\
\bottomrule
\end{tabular}%
\end{adjustbox}
\end{table*}
Comparing the results presented in~\cref{tab:limited_search_space} with the performances in~\cref{tab:downstream_task}, we beat the Naive (R) experiments (row 5 in~\cref{tab:downstream_task}) in all cases for the PneumoniaMNIST and 80\% of the time for BreastMNIST and AIforCOVID.
We observed a lack of direct proportionality between the g-mean on the real test set and $|G|_M$, which underlines two main findings.
First, the procedure is robust to the initial search space. Indeed, it defines the best achievable GAN ensemble independently from the GAN models.  
Second, the results are favourable due to sampling GANs that did not fail during training. 
For example, analysing the set of models randomly selected for PneumoniaMNIST when $|G|_M$ is equal to 7 (last row in~\cref{tab:limited_search_space}), we sampled ACGAN-Mod-TAC and ReACGAN-TAC that achieved the lowest performances on the test set as reported in~\cref{tab:single_gan_XRay}.
Including more training iterations further alleviates this dependency as shown in~\cref{tab:single_gan_XRay,tab:single_gan_ultrasound,tab:single_gan_aiforcovid}.

\subsection{Battle of the Backbones for GAN Embeddings}
\label{sec:bob}
In this section, we study the role of different feature extractors, beyond the conventional ImageNet pre-trained models, in evaluating synthetic images from generative models.
The reliance on ImageNet features often introduces bias, particularly in fields like medical imaging, where data distributions may differ from that of natural images as in, ImageNet~\cite{kynkaanniemi2022role}.
To address this, we explore a variety of \textit{backbones}, including different architectures, domain-specific fine-tuning, and unsupervised learning models.
Our investigation encompasses five experimental setups. 
These experiments span supervised models (InceptionV3 and ResNet50), their medical domain fine-tuned counterparts (InceptionV3-Med and ResNet50-Med), and a self-supervised model (SwAV)~\cite{caron2020unsupervised}.
We summarise the backbones configuration in~\cref{tab:backbones} in~\cref{app:Backbones_analysis_details}.
Moreover, in~\cref{app:Backbones_analysis_details}, we report additional details about the backbone architectures and training procedures.
In~\cref{tab:backbones_analysis}, we underline and report in bold the worst and best results, respectively.

We present the results of the current analysis in~\cref{tab:backbones_analysis}, where we compare the best possible solution achievable running the Pareto-optimisation to select the ensemble $G^*$, i.e., Oracle, but changing backbones.
We calculate the g-mean on the real test set and the gap, $\gamma_{SS}$.
This gap denotes the performance difference when training the ResNet18 with synthetic data from ensembles selected using backbones different from SwAV for the embedding extraction step.
Moreover, we include the number of GANs selected by each solution.
\begin{table*}[tb]
\caption{Performances on the real test set when training the downstream model with $G^{*}$ selected using different backbones.
$\gamma_{SS}$ represents the \% g-mean gap in training the ResNet18 with $G^{*}$ selected according to the SwAV's embedding compared to others.
The best and worst solutions are shown in bold and underlined, respectively.
}
\label{tab:backbones_analysis}
\centering
\begin{adjustbox}{width=\textwidth}
\begin{tabular}{lccccccccc}
\toprule
                   & \multicolumn{3}{c}{\textbf{PneumoniaMNIST}} & \multicolumn{3}{c}{\textbf{BreastMNIST}}   & \multicolumn{3}{c}{\textbf{AIforCOVID}}     \\
\textbf{Backbones} & g-mean $\uparrow$         & $\gamma_{SS}$ $\downarrow$  & $\#G_i$ $\downarrow$ & g-mean $\uparrow$        & $\gamma_{SS}$ $\downarrow$   & $\#G_i$ $\downarrow$ & g-mean  $\uparrow$       & $\gamma_{SS}$ $\downarrow$ & $\#G_i$ $\downarrow$ \\
\midrule
SwAV (baseline)               &  \bm{$0.881$}     & --              & 40      & \bm{$0.755$}    & --              & 38      & 0.573          & --               & 30      \\
\hline
InceptionV3        & 0.879           & $-0.2$          & 39      & 0.738          & $-2.2$          & 37      & \bm{$0.612$}    & \bm{$+6.8$}     & 35      \\
ResNet50           & 0.874           & $-0.8$          & 35      & 0.733          & $-2.9$          & 34      & 0.563          & $-1.7$           & 26      \\
\hline
InceptionV3-Med    & 0.877           & $-0.5$          & 42      & {\ul 0.684} & {\ul $-9.4$} & 32      &{\ul 0.497} & {\ul $-13.3$} & 33      \\
ResNet50-Med       & {\ul 0.873}  & {\ul $-0.9$} & 37      & 0.705          & $-6.6$          & 44      & 0.566          & $-1.2$           & 38      \\
\bottomrule
\end{tabular}%
\end{adjustbox}
\end{table*}
Row one in~\cref{tab:backbones_analysis} shows the results achieved when finding $G^*$ using SwAV, which is the baseline approach.
Focusing on rows two and three in~\cref{tab:backbones_analysis}, InceptionV3 and ResNet50 demonstrated competitive yet inconsistent results across tasks. 
InceptionV3's performance for the AIforCOVID task surpasses SwAV's, highlighting the potential of using pre-trained models in specific scenarios. 
The finding suggests that supervised-driven embeddings, i.e., representations extracted from models pre-trained via supervised learning, are still effective in certain but not all scenarios. 
Overall, the InceptionV3 and ResNet50 embeddings allow using fewer GANs and lower values of $\#G_i$, with respect to SwAV-based selection.
Rows four and five in~\cref{tab:backbones_analysis} show that fine-tuning each backbone using domain-specific data (medical images) did not improve performance.
To fine-tune each backbone, we used each dataset's training set.
Both InceptionV3-Med and ResNet50-Med show a decrease in g-mean, especially in BreastMNIST and AIforCOVID.
This might indicate that domain-specific fine-tuning can lead to overfitting or loss of generalisability.
The SwAV outperformed the other models across all tasks, with the highest scores in the PneumoniaMNIST and BreastMNIST datasets.
These findings underscore the complexities in backbone selection for GAN evaluation in medical imaging, suggesting a preference for self-supervised models like SwAV for their robustness and generalisability, as seen elsewhere in the literature~\cite{kynkaanniemi2022role,goldblum2023battle}.

\subsection{Fidelity, Diversity, and Utility Analysis}
\label{sec:fidelity_diversity}

We evaluated the quality of synthetic data generated by various GANs compared to our solution, $G^{*}$, in terms of fidelity and diversity.
\begin{table*}[tb]
\caption{Metrics to evaluate the quality of generated images. We report the best results for each metric in bold. 
*Results obtained ranking GAN’s according to the downstream task performances on the real test set.}
\label{tab:metrics_analysis}
\centering
\begin{adjustbox}{width=\textwidth}
\begin{tabular}{lccccccccc}
\toprule
                 & \multicolumn{3}{c}{\textbf{PneumoniaMNIST}}      & \multicolumn{3}{c}{\textbf{BreastMNIST}}         & \multicolumn{3}{c}{\textbf{AIforCOVID}}    \\
\textbf{}        & FID $\downarrow$           & Density $\uparrow$       & Coverage $\uparrow$      & FID  $\downarrow$          & Density $\uparrow$       & Coverage $\uparrow$       & FID $\downarrow$            & Density $\uparrow$        & Coverage $\uparrow$ \\
\midrule
Single (top-1)$^*$  & 2.604          & 0.232          & 0.292          & 4.551          & 0.692          & 0.769          & \textbf{6.477} & \textbf{0.031} & 0.071    \\
Single (average) & 6.020          & 0.721          & 0.460          & 15.621         & 0.206          & 0.240          & 19.751         & 0.002          & 0.000    \\
\hline
Naive (R)        & 1.643          & 0.781          & 0.999          & 6.279          & 0.482          & 0.978          & 13.818         & 0.002          & 0.002    \\
Naive (A)        & 1.880          & 0.721          & \textbf{1.000} & 4.759          & 0.546          & \textbf{1.000}          & 11.437         & 0.002          &   \textbf{0.122}    \\
\hline
$G^*$ (our)       & \textbf{0.915} & \textbf{0.886} & \textbf{1.000} & \textbf{3.900}          & \textbf{0.780}          & \textbf{1.000} & 11.683         & 0.005          & 0.117    \\
\bottomrule
\end{tabular}%
\end{adjustbox}
\end{table*}

In~\cref{tab:metrics_analysis}, we report the Frechet Inception Distance (FID), density, and coverage for the synthetic data created with Single (top-1), Single (average), Naive (A), Naive (R), and $G^{*}$ approaches.
In the first and second rows of~\cref{tab:metrics_analysis} are the results for the Single (top-1) and Single (average) experiments. 
In rows three and four of~\cref{tab:metrics_analysis} are the results for the Naive (R) and Naive (A) experiments.
The performances of the proposed approach, $G^*$, are in the last row.
Note that density and coverage independently evaluate fidelity and diversity, but FID does not differentiate between them.
For the BreastMNIST and AIforCOVID datasets, the Single (average) experiment performs worse (higher FID, lower density and coverage) compared to the Single (top-1).
However, for the PneumoniaMNIST dataset, there is no clear consensus between the FID, density, and coverage metrics.
To further validate the proposed multi-objective optimisation when varying the distribution fidelity metric $d$, we performed the ablation test in~\cref{app:ablation_test}.
The results for the Naive experiments are in the third and fourth rows on~\cref{tab:metrics_analysis}. 
They show increased coverage, implying that assembling different $G_i$ may increase diversity.
Randomly selecting the GANs or naively including them all decreases the fidelity.
Comparing the Single (average) and Naive (A) experiments, we observe an increased diversity while having equal (PneumoniaMNIST) or lower fidelity (BreastMNIST and AIforCOVID). 
The last row of~\cref{tab:metrics_analysis} shows that the proposed solution allowed us to increase the diversity to Single GAN while keeping fidelity. 
Specifically, we found lower FID and higher diversity and fidelity for PneumoniaMNIST and BreastMNIST.
However, the AIforCOVID dataset exhibits a decrease in fidelity yet an increase in diversity compared to the Single (top-1) experiment. 
AIforCOVID is the only dataset where the top-1 GAN surpasses the real test set g-mean performance achieved by the proposed solution~\cite{bib:ruffini2024multi,bib:guarrasi2024multimodal}, as discussed in~\cref{sec:Impact_downstream}.

\begin{figure*}[tb]
  \centering
  \begin{subfigure}{0.31\linewidth}
    \includegraphics[width=1.0\linewidth]{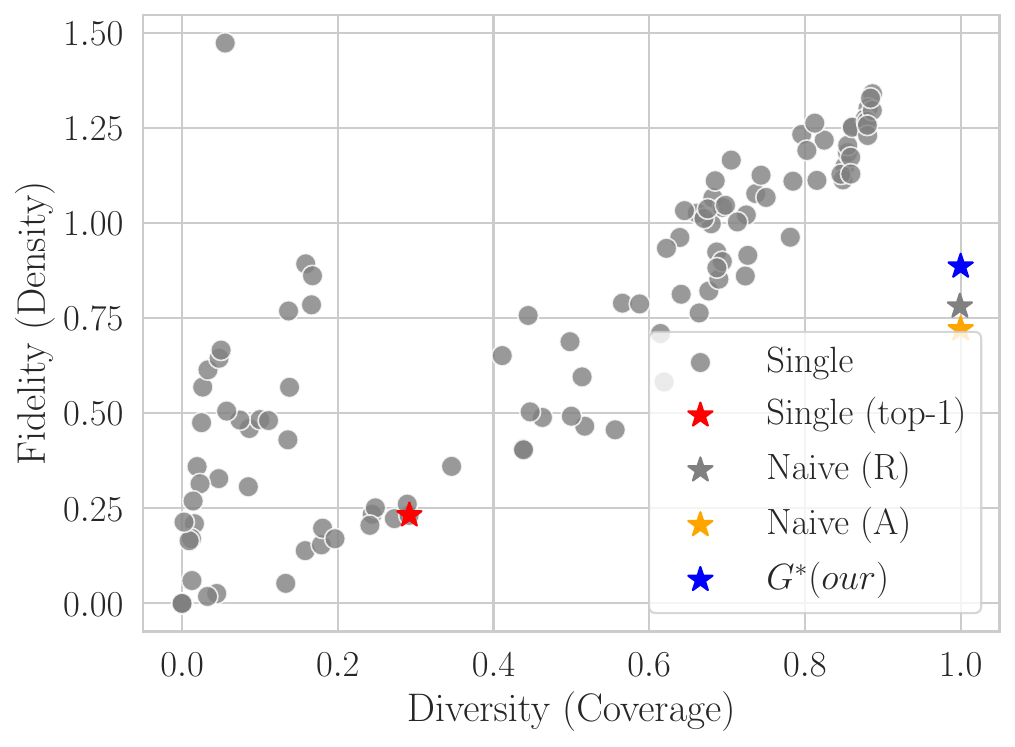}
    \caption{PneumoniaMNIST.}
   \label{fig:diversity_fidelity_XRay}
  \end{subfigure}
  \hfill
  \begin{subfigure}{0.31\linewidth}
    \includegraphics[width=1.0\linewidth]{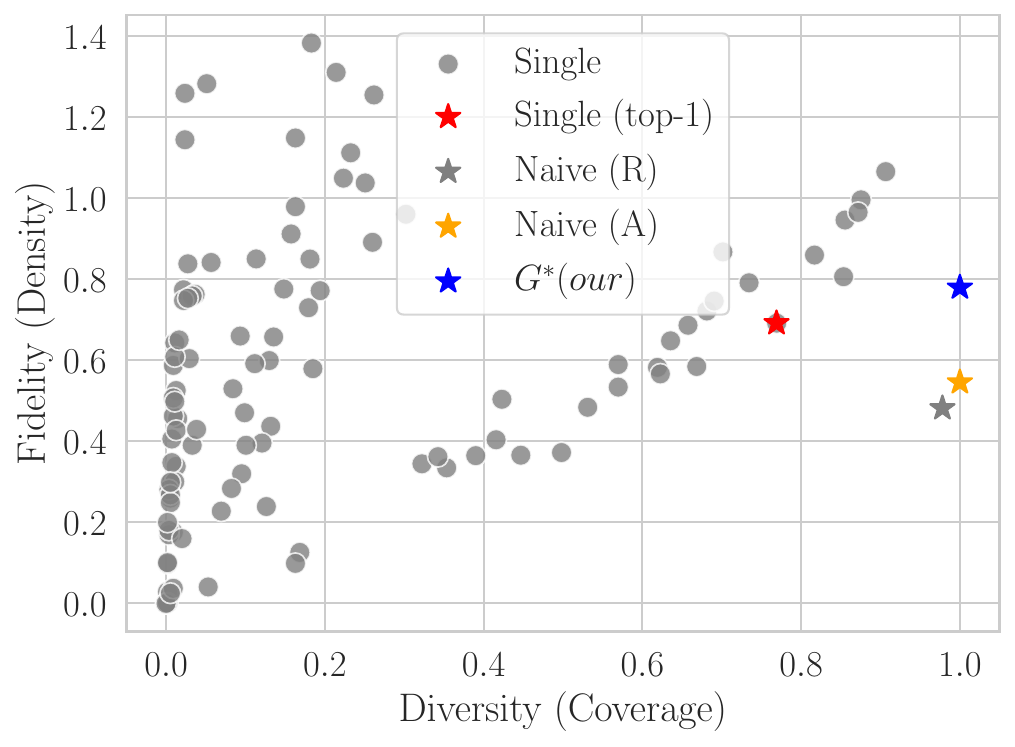}
    \caption{BreastMNIST.}
    \label{fig:diversity_fidelity_ultrasound}
  \end{subfigure}
\hfill
    \begin{subfigure}{0.31\linewidth}
    \includegraphics[width=1.0\linewidth]{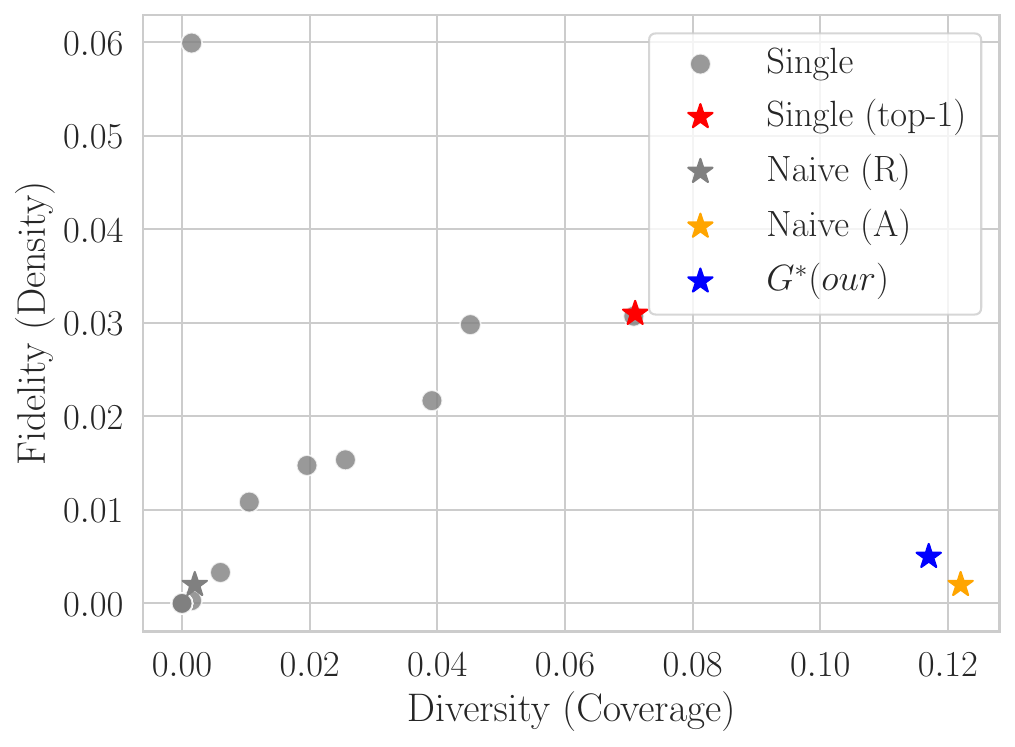}
    \caption{AIforCOVID.}
    \label{fig:diversity_fidelity_XRayhq}
  \end{subfigure}

    \begin{subfigure}{0.31\linewidth}
    \includegraphics[width=1.0\linewidth]{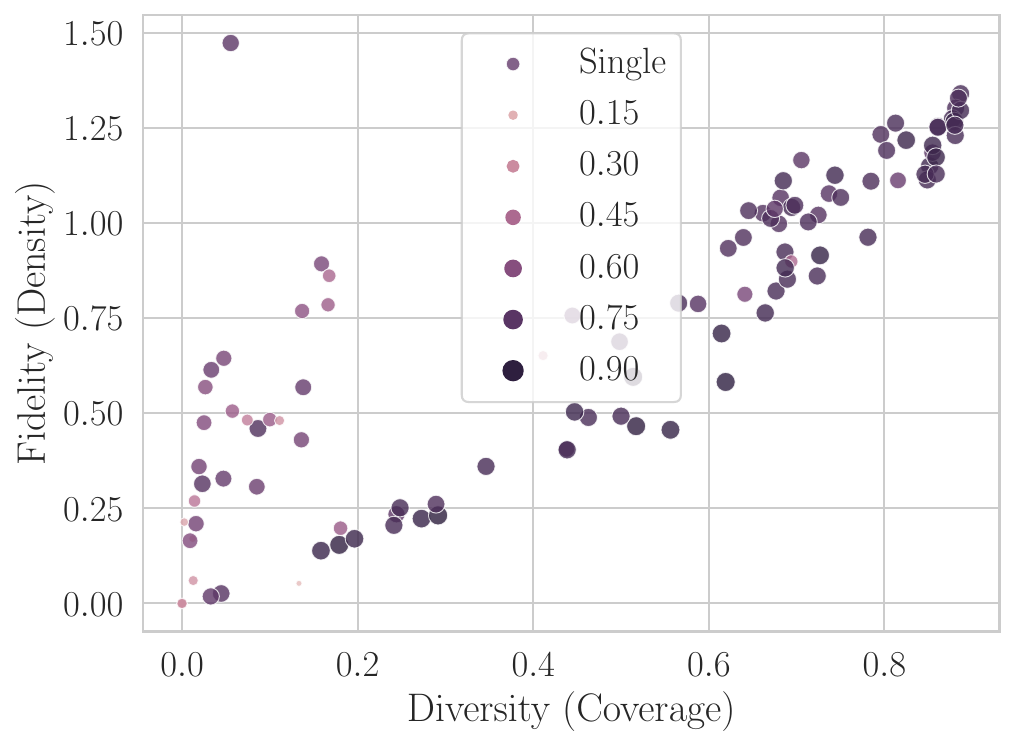}
    \caption{PneumoniaMNIST.}
   \label{fig:diversity_fidelity_g_mean_XRay}
  \end{subfigure}
  \hfill
  \begin{subfigure}{0.31\linewidth}
    \includegraphics[width=1.0\linewidth]{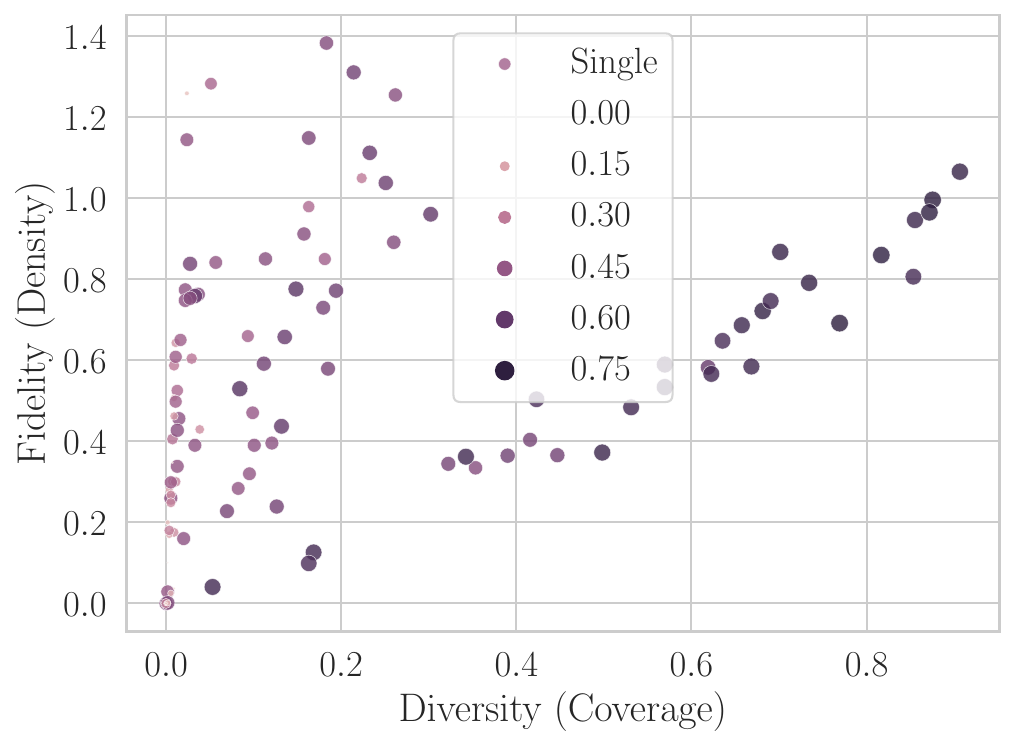}
    \caption{BreastMNIST.}
    \label{fig:diversity_fidelity_g_mean_ultrasound}
  \end{subfigure}
\hfill
    \begin{subfigure}{0.31\linewidth}
    \includegraphics[width=1.0\linewidth]{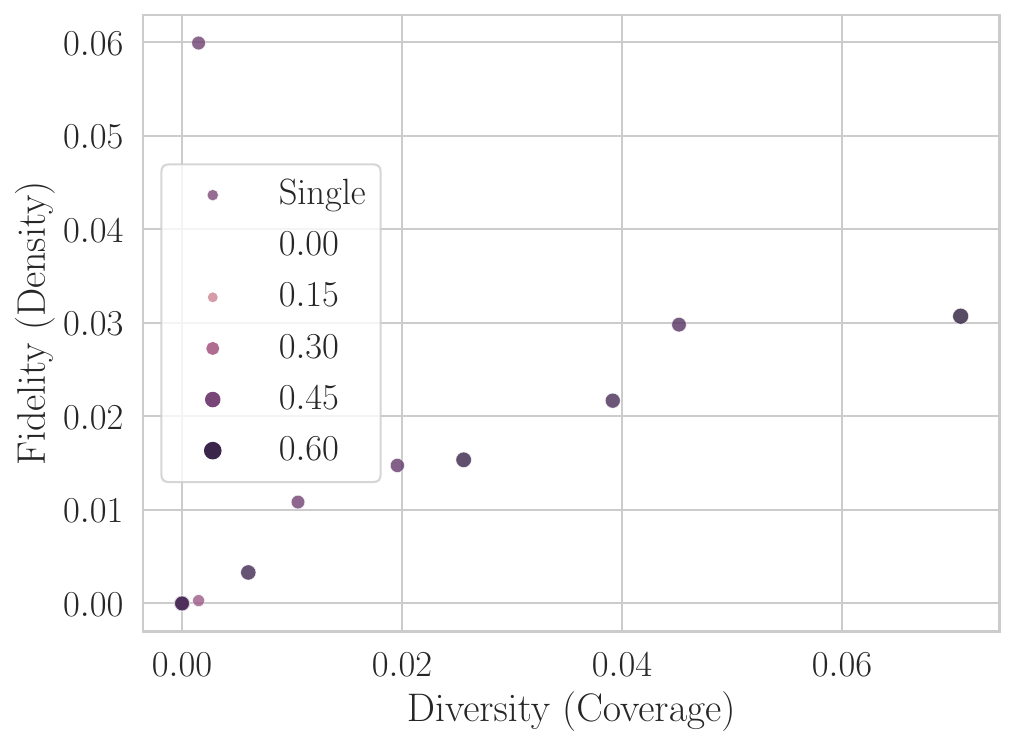}
    \caption{AIforCOVID.}
    \label{fig:diversity_fidelity_g_mean_XRayhq}
  \end{subfigure}
  \caption{Plots of Diversity, Fidelity, and Utility using SwAV embeddings. 
  The top row is single GANs (grey circles), Naive (A) and Naive (R) (grey and orange star), and the optimal ensemble $G^{*}$ (blue star). 
  The bottom row shows the real test set downstream performances using the synthetic data from single GANs.
  Colours near pink or purple and circles of lower or higher diameters indicate lower or higher test set performances.}
  \label{fig:diversity_fidelity_analysis}
\end{figure*}

\cref{fig:diversity_fidelity_analysis} presents an in-depth analysis of the performances of the single GAN models.
Each subplot reports the diversity (first axis) versus fidelity (second axis) using the notion of coverage and density metrics~\cite{naeem2020reliable}. 
The top row,~\cref{fig:diversity_fidelity_XRay}, \ref{fig:diversity_fidelity_ultrasound}, \ref{fig:diversity_fidelity_XRayhq}, compares the fidelity and diversity of Single and Naive experiments against the ensemble approaches.
The bottom row,~\cref{fig:diversity_fidelity_g_mean_XRay}, \ref{fig:diversity_fidelity_g_mean_ultrasound}, \ref{fig:diversity_fidelity_g_mean_XRayhq}, shows how fidelity and diversity impact the synthetic data utility for the downstream task.
Focusing on the top row, a red star means the Single (top-1) experiment, yellow and grey stars mean the Naive (A) and Naive (R) experiments and a blue star means proposed $G^*$.
In~\cref{fig:diversity_fidelity_XRay}, \ref{fig:diversity_fidelity_ultrasound}, \ref{fig:diversity_fidelity_XRayhq}, the proposed approach (blue star) demonstrates better diversity compared to individual $G_i$ (grey circles), indicating an improved mode coverage.
The yellow and grey stars indicate a similar diversity compared to $G^{*}$ for two datasets out of three, the PneumoniaMNIST (\cref{fig:diversity_fidelity_XRay}) and BreastMNIST (\cref{fig:diversity_fidelity_ultrasound}) but reduced fidelity in all cases.
This may depend on the fidelity of the single GANs.
In this setting, the number of training failures is higher, reflected in lower fidelity when using the Naive (R) and Naive (A) approaches.

The bottom row (\cref{fig:diversity_fidelity_g_mean_XRay}, \ref{fig:diversity_fidelity_g_mean_ultrasound}, \ref{fig:diversity_fidelity_g_mean_XRayhq}) shows that diversity is more crucial to ensure synthetic data utility for downstream applications.
After a certain diversity threshold on the first axis, we see that the synthetic data from $G_i$ achieve similar real test-set performances. 
Whereas having high fidelity, i.e., high values on the second axis, does not guarantee good downstream task performances.

\subsection{Clinical Impact}

The proposed study has significant implications for clinical practice and research by addressing the persistent challenges in synthetic medical image generation—namely fidelity, diversity, and efficiency. By leveraging an optimally selected ensemble of GANs, this work enables the generation of high-quality synthetic medical images that accurately reflect the underlying data distribution. These synthetic datasets can mitigate the limitations of real-world medical data, such as scarcity, imbalance, and privacy concerns, thereby enhancing the robustness and generalizability of diagnostic and predictive models.

Moreover, the improved fidelity and diversity of the generated images provide a valuable resource for training and validating AI-based diagnostic tools. This is particularly critical for rare diseases or conditions where large, balanced datasets are often unavailable. The computational efficiency achieved through the proposed ensemble approach also facilitates the rapid generation of synthetic data, making it feasible for integration into routine workflows, such as clinical decision support, imaging protocol development, and educational purposes.

By ensuring minimal redundancy and maximizing utility across models in the ensemble, this approach reduces the risk of overfitting and enhances the transferability of AI models to new clinical settings. Additionally, the comprehensive evaluation across multiple medical datasets and the use of diverse GAN architectures make the findings broadly applicable across imaging modalities. This has the potential to accelerate innovation in personalized medicine, improve diagnostic accuracy, and reduce healthcare costs by enabling more efficient and equitable access to advanced AI-driven imaging solutions.

\section{Conclusions} \label{sec:conclusion}
GANs struggle with the generative learning trilemma, achieving high-quality and fast sampling but often showing low sampling diversity, limiting their utility in real-world applications. 
We propose a method to ensemble GANs that improves synthetic data diversity and fidelity, crucial for generating accurate medical imaging data. Our approach uses multi-objective optimisation to select a Pareto-optimal set of GANs, maximising the coverage of diverse medical conditions and anomalies.
Extensive experiments demonstrate that synthetic datasets generated from such an ensemble improve downstream task performance compared to single GANs and naive selection approaches, specifically enhancing diagnostic modelling in medical applications. We also analysed how increasing fidelity and diversity in synthetic medical images impacts downstream task utility. The proposed method can be applied to any GAN search space, providing an optimal non-overlapping model combination that maximises fidelity and diversity, relieving analysts from deciding \textit{which} GAN architectures are best for medical imaging or \textit{when} to stop training the GAN.

A key limitation is the static sampling from each GAN in the ensemble during downstream model training. This means once the GANs in the ensemble are selected, their relative contribution is fixed, not accounting for the possibility that some GANs are more relevant for the downstream tasks or that certain synthetic samples are more beneficial at different training stages. Future work will focus on creating a dynamic ensemble that adapts based on feedback from the downstream model. 
Additionally, the current methodology requires training numerous networks before selecting the optimal ensemble, which can be computationally expensive, particularly with high-resolution medical data.
To address this, a future direction includes defining an overall computational budget encompassing GAN training, ensemble search, and downstream tasks. This approach would optimise the entire process within a fixed budget, balancing the computational resources between training and optimisation to take steps for downstream tasks concurrently, thus reducing the overall computational burden and offering a more feasible solution for medical research and application.

\section*{Acknowledgment}
This work was partially founded by: 
i) Università Campus Bio-Medico di Roma under the program ``University Strategic Projects'' within the project ``AI-powered Digital Twin for next-generation lung cancEr cAre (IDEA)''; 
ii) from PRIN 2022 MUR 20228MZFAA-AIDA (CUP C53D23003620008); 
iii) from PRIN PNRR 2022 MUR P2022P3CXJ-PICTURE (CUP C53D23009280001);
iv) from PNRR MUR project PE0000013-FAIR.

\bibliographystyle{elsarticle-num-names} 
\bibliography{main_etal.bib}

\renewcommand{\thesection}{A\arabic{section}}
\renewcommand{\thetable}{A\arabic{table}}
\renewcommand{\thefigure}{A\arabic{figure}}

\section*{\centering Appendix}

\setcounter{section}{0}
\setcounter{table}{0}
\setcounter{figure}{0}

\section{Downstream model setups}
\label{app:training_evaluation_setup}
We performed different downstream tasks according to the different datasets, and presented in~\cref{app:Datasets}.
We utilised a ResNet18 model as the downstream model in our experiments. 
This model is characterised by fewer trainable parameters than more complex models.
We employed Adam optimiser to train the model, using a batch size of 64 and an initial learning rate of 0.001.
We used a learning rate scheduler that reduced the learning rate by a factor of ten if the validation loss did not improve for five epochs.
With an early stopping criteria, we stopped the training if no improvement was seen for 25 epochs.
Before early stopping, we employed a warm-up period of 25 epochs.
We set the maximum number of epochs to 100.
For preprocessing, the images underwent the same steps as those used for training the GANs mentioned above, ensuring consistency, i.e., resize using Lanczos filter for interpolation, but normalising to $[0,1]$.
Both the validation and test sets contained only real images.
Each ensemble experiment was conducted with a training set for the ResNet18 network that had the same size as the real training dataset.
This approach ensured uniformity across different experimental setups. 
Finally, to provide reliable performance metrics and to reduce the variability inherent in model training, the results were averaged over 20 training runs.

\section{Backbone analysis}
\label{app:Backbones_analysis}
We define as a \textit{backbone} the model used to extract features from real and synthetic data for evaluating generative models.
We explored the impact on downstream task performance when using different backbone architectures to search for the optimal ensemble $G^*$.
We tested variations in ImageNet-pre-trained models, i.e., different architectures,
domain-specific fine-tuning, and unsupervised learning approaches. 
To this scope, we performed a total of five experiments, summarised in~\cref{tab:backbones}.
\begin{table}[h]
\caption{The backbones we benchmarked.}
\label{tab:backbones}
\centering
\begin{tabular}{lccc}
\toprule
{\textbf{Backbones}}      & {\textbf{Supervision}} & {\textbf{Fine tuning}} & \textbf{{Architecture}} \\
\midrule
SwAV (baseline)          & \xmark   & \xmark           & ResNet50     \\ \midrule

InceptionV3     & \cmark     & \xmark           & InceptionV3  \\
ResNet50        & \cmark     & \xmark           & ResNet50     \\
\midrule
InceptionV3-Med & \cmark     & \cmark          & InceptionV3  \\
ResNet50-Med    & \cmark     & \cmark         & ResNet50     \\
\bottomrule
\end{tabular}%
\end{table}

In the following, we report the training and evaluation protocols for backbones and some additional results for the Pareto optimisation procedure.

\subsection{Training and evaluation setups}
\label{app:Backbones_analysis_details}

For the fine-tuning of InceptionV3 and ResNet50, i.e., InceptionV3-Med and ResNet50-Med in~\cref{tab:backbones},
we converted grayscale images to three channels by repetition, and we resized to $299 \times 299$ for InceptionV3 and $224 \times 224$ for ResNet50.
We resized both real and synthetic data using bilinear interpolation (using Pillow).
We used the same interpolation adopted in the ImageNet-pretraining procedure of each backbone to avoid possible bias due to preprocessing inconsistencies.
Then, the images were normalised to $[0,1]$ and standardised as per ImageNet's common mean and standard deviation.
We train each network using the Adam optimiser, with a batch size of 64 and an initial learning rate of $0.001$. 
We implemented a scheduler that reduced the learning rate by a factor of ten if there was no improvement in validation loss for ten consecutive epochs.
We used early stopping after 20 epochs without validation loss improvement after a warm-up phase of 20 epochs.
The maximum number of epochs for training was set to 100.
We fine-tuned the networks separately for each dataset and for the task at hand.

For computing the distribution quality metric $d$, i.e., FID or harmonic mean of density and coverage, we compared the same number of synthetic and real training images. 
During inference, real and synthetic images were fed to each backbone and underwent the same preprocessing procedure explained above for fine-tuning, i.e., conversion to three channels, resize, normalization to $[0,1]$, and standardization.

\subsection{Analysis of Pareto Plots}
\label{app:Backbones_analysis_pareto}

\begin{figure*}[tb]
  \centering
  \begin{subfigure}{0.31\linewidth}
   \includegraphics[width=1.0\linewidth]{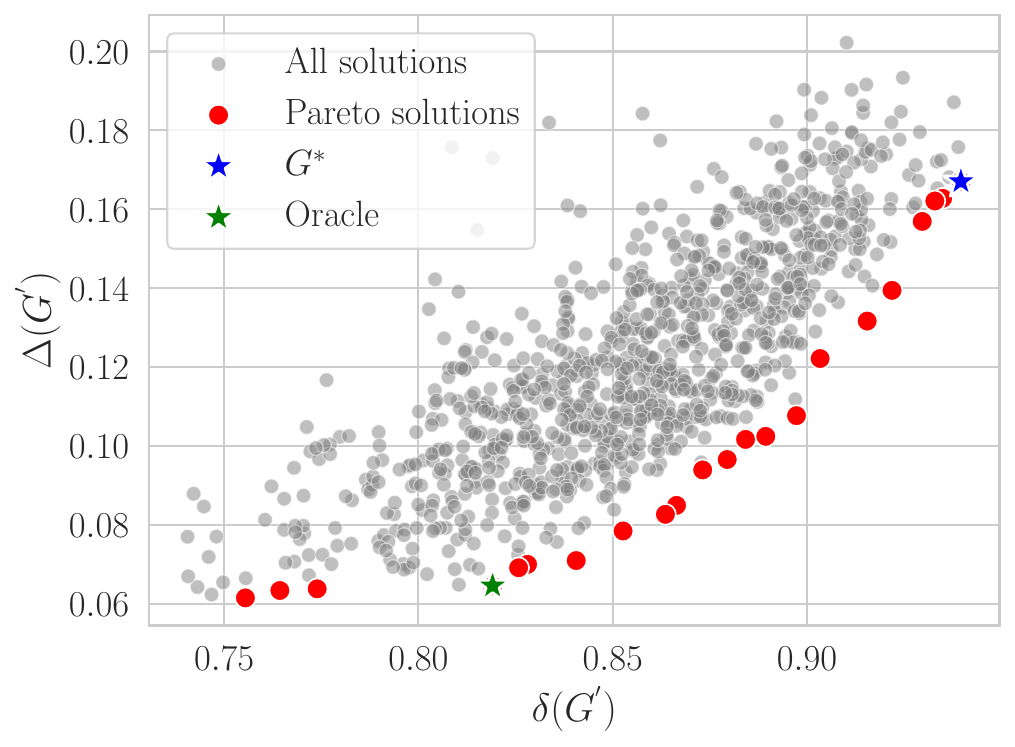}
   \caption{PneumoniaMNIST.}
   \label{fig:pareto_plots_swav_XRay}
  \end{subfigure}
  \hfill
  \begin{subfigure}{0.31\linewidth}
   \includegraphics[width=1.0\linewidth]{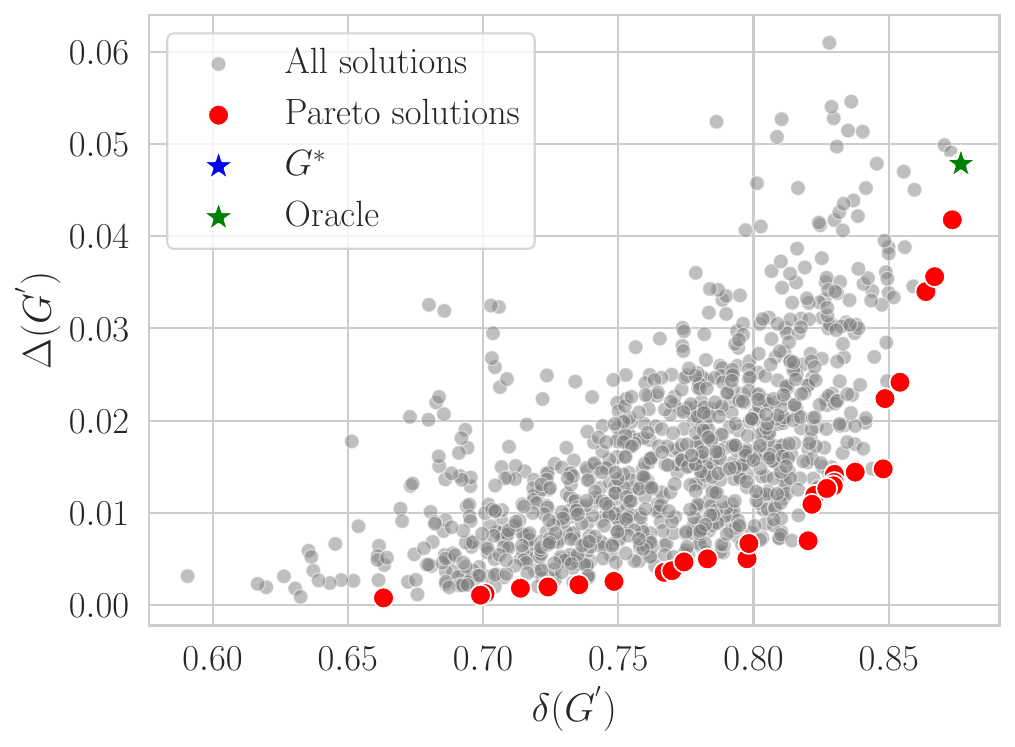}
   \caption{BreastMNIST.}
   \label{fig:pareto_plots_swav_ultrasound}
  \end{subfigure}
  \hfill
  \begin{subfigure}{0.31\linewidth}
   \includegraphics[width=1.0\linewidth]{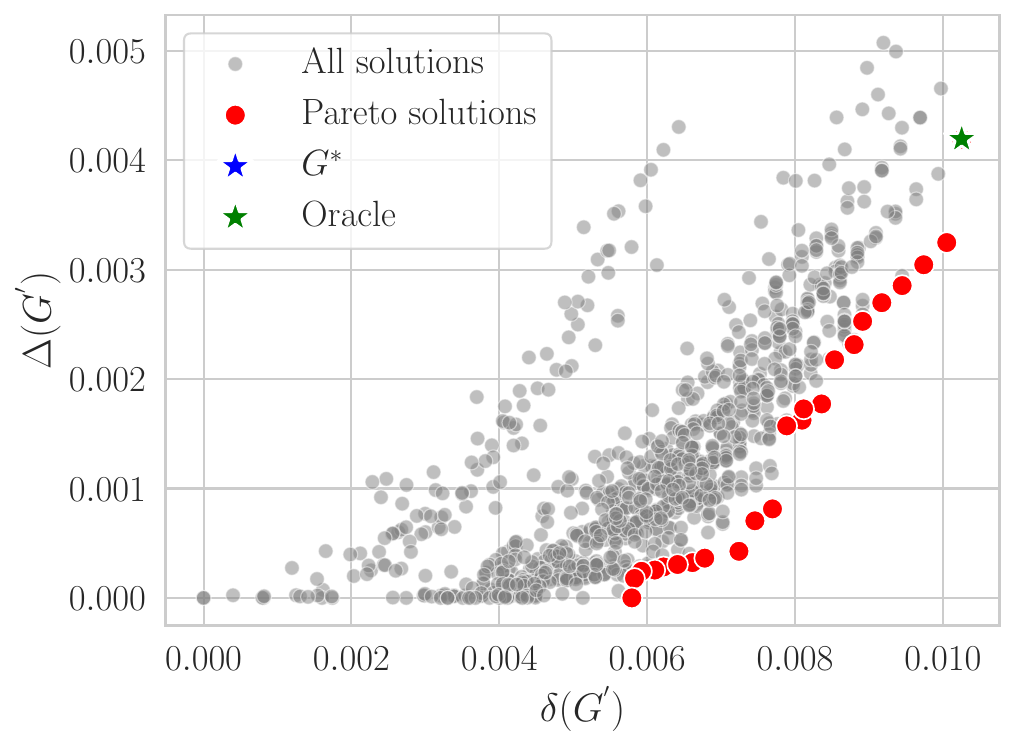}
   \caption{AIforCOVID.}
   \label{fig:pareto_plots_swav_AIforCOVID}
  \end{subfigure}
  \caption{Pareto plots for each dataset using SwAV as a backbone.}
  \label{fig:pareto_plots_swav}
\end{figure*}
In the context of our proposed ensemble GAN method, a key aspect of our methodology is the optimisation process, described in~\cref{sec:methods}.
To find the optimal ensemble, we run the optimisation over 1\,000 iterations. 
\cref{fig:pareto_plots_swav} illustrates the Pareto plots for SwAV, i.e., the outcomes of our multi-objective optimisation when using the embeddings extracted using SwAV.
The plot visualises all 1\,000 solutions (represented as grey circles) and highlights the Pareto optimal solutions (denoted as red circles) that fulfil the inequalities~\cref{eq:pareto_condition}. 
A key observation from~\cref{fig:pareto_plots_swav_XRay}, \ref{fig:pareto_plots_swav_ultrasound}, \ref{fig:pareto_plots_swav_AIforCOVID} is the alignment of our Pareto optimal solutions with the Oracle in 2 out of 3 datasets when the selection criterion minimises Intra-d (as specified in~\cref{eq:selection_criteria}). 

\begin{figure}[tb]
  \centering
   \includegraphics[width=0.7\linewidth]{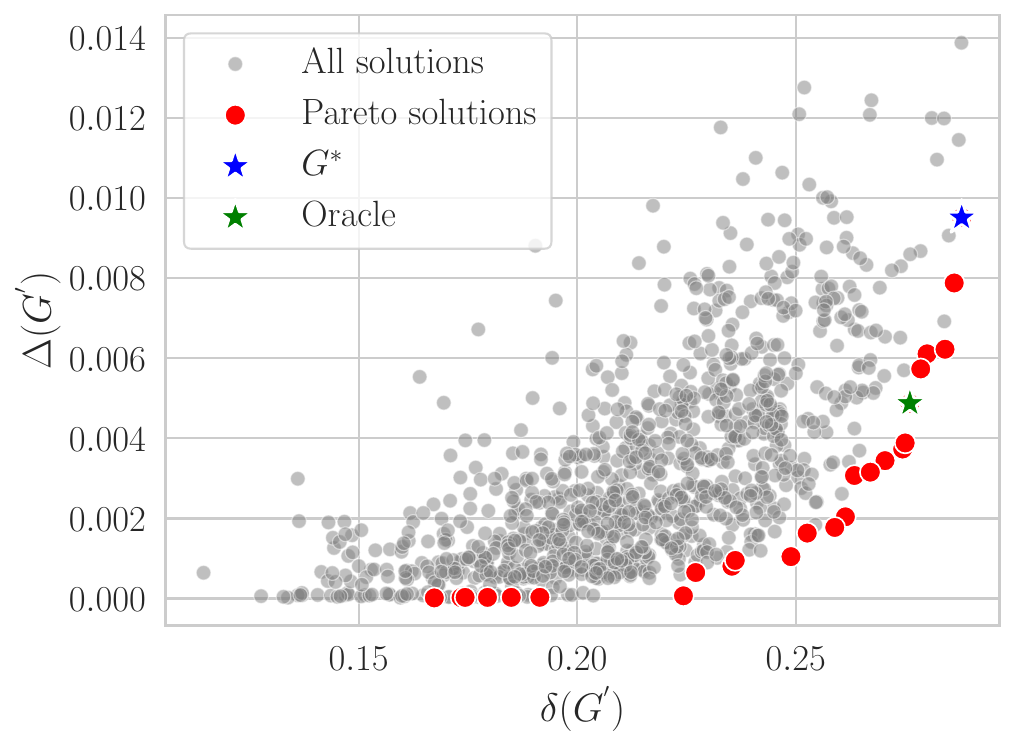}
   \caption{Pareto plots for AIforCOVID using InceptionV3 as a backbone.}
   \label{fig:pareto_plots_inceptionv3_aiforcovid}
\end{figure}

The Pareto solutions for 
for the AIforCOVID dataset shows small differences within the range of $\delta(G^{'})$ and $\Delta(G^{'})$, meaning that different ensemble solutions from the SwAV's Pareto plot achieve a similar Intra-$d$ and Inter-$d$.
The low variability between different solutions suggests that SwAV does not distinguish among different ensemble solutions.
In this scenario, a subtle difference exists between performing the ensemble optimisation or only randomly picking a set of GANs, as shown by the inferior ResNet18 performances for AIforCOVID, when using SwAV to select $G^*$ (first first-row of~\cref{tab:backbones_analysis} in the AIforCOVID column).
Conversely, InceptionV3 can extract more informative embeddings than SwAV, as shown by the Pareto frontier in~\cref{fig:pareto_plots_inceptionv3_aiforcovid}, which shows a sparser set of solutions on the Pareto frontier, effectively discriminating between the different ensembles tested during the optimisation.
This resulted in improved performance in the downstream task as outlined in the second-row of~\cref{tab:backbones_analysis} in the AIforCOVID column.

\section{Ablation test}
\label{app:ablation_test}
\begin{table}[tb]
\caption{Ablation test.
The first two rows and the last two rows show the results for the multi- and uni-objective approaches, respectively, when using dns,cvg and FID.
We report in bold the results where the multi-objective approach outperforms the uni-objective approach.}
\label{tab:ablation}
\centering
\begin{adjustbox}{width=\columnwidth}
\begin{tabular}{cccccccc}
\toprule
\multirow{2}{*}{\textbf{Criteria}} & \multirow{2}{*}{\textbf{Metric} \bm{$d$}} & \multicolumn{2}{c}{\textbf{PneumoniaMNIST}} & \multicolumn{2}{c}{\textbf{BreastMNIST}} & \multicolumn{2}{c}{\textbf{AIforCOVID}} \\
 &  & g-mean $\uparrow$ & $\#G_i$   $\downarrow$ & g-mean $\uparrow$ & $\#G_i$   $\downarrow$ & g-mean $\uparrow$ & $\#G_i$   $\downarrow$ \\ \midrule
\multirow{2}{*}{Intra-$d$ $\uparrow$, Inter-$d$ $\downarrow$} & dns, cvg & \bm{$0.867$} & 38 & \bm{$0.755$} & 32 & \bm{$0.573$} & 30 \\
 & FID & \bm{$0.875$} & 42 & \bm{$0.760$} & 45 & \bm{$0.554$} & 48 \\ \hline
\multirow{2}{*}{Intra-$d$ $\uparrow$} & dns, cvg & $0.861$ & $41$ & $0.716 $ & $44$ & $0.528$ & $31$ \\
 & FID & $0.860$ & $55$ & $0.736$ & $57$ & $0.526$ & $46$ \\ 
 \bottomrule
\end{tabular}%
\end{adjustbox}
\end{table}
          
To further validate the proposed multi-objective optimisation, we performed an ablation test.
We shifted to a uni-objective approach and focused only on maximising Intra-$d$ (Intra-$d$ $\uparrow$) instead of maximising Intra-$d$ and minimising Inter-$d$ (Intra-$d$ $\uparrow$, Inter-$d$ $\downarrow$).
Moreover, we assessed the reliability of the multi-objective optimisation when changing the distribution quality metric, $d$. 
For this purpose, we tested the Fréchet Inception Distance (FID) as the standard metric used in the literature to assess the generation quality of generative model~\cite{heusel2017gans}.
This test was crucial to determine whether maximising Intra-$d$ and minimising Inter-$d$ make a difference in finding $G^{*}$ to optimising only the fidelity term Intra-$d$.
We summarised the results in~\cref{tab:ablation}, where we reported the real test set g-mean and the number of $G_i$ in the ensemble. 
The first two rows show the results when using the multi-objective optimisation approach for dns,cvg and FID, while the third and fourth rows show the results achieved ablating the Inter-$d$ terms.
Focusing on the first and second-to-last rows (dns,cvg experiments) and the second and last rows (FID experiments) in~\cref{tab:ablation}, we found that for all datasets the synthetic data from the ensemble selected with the multi-objective optimisation achieved better real test set g-mean and using fewer $G_i$. 
Thus, adding the Inter-$d$ terms in the final objective has the effect of discharging overlapping GANs, i.e., models that generate similar synthetic data.
This has the effect of reducing similar data in training the downstream model, reducing the overfitting.
Moreover, the analysis showed that the proposed approach is robust to different distribution quality metrics, $d$.

The key takeaway from this analysis is that ensembling multiple synthetic datasets from different GANs can improve downstream performance on real test data.
However, implementing selection criteria for naive selection is crucial.
In particular when there is a high risk of failed GAN training.

\section{Single GAN analysis: which and when} 
\label{app:single_gan_analysis}

\begin{table}[tb]
\caption{Single GAN g-mean scores for PneumoniaMNIST dataset.
We report in bold the experiments where Naive (I) or Naive (M) are above the mean performances of models (Mean (M)) and iterations (Mean (I)), respectively.
We underline the best overall experiment. We highlight in green and yellow the top-1 and top-5 performing GANs, respectively. We denote with * the GANs selected by our multi-objective optimisation.}
\label{tab:single_gan_XRay}
\centering
\begin{adjustbox}{width=\columnwidth}
\begin{tabular}{l lllll ll}
\toprule
\textbf{\diagbox[linewidth=0pt]{Models}{Iterations}} & \multicolumn{1}{c}{\textbf{20000}} & \multicolumn{1}{c}{\textbf{40000}} & \multicolumn{1}{c}{\textbf{60000}} & \multicolumn{1}{c}{\textbf{80000}} & \multicolumn{1}{c}{\textbf{100000}} & \textbf{Mean (I)} & \textbf{Naive (I)} \\
\midrule
\textbf{ACGAN-Mod} & 0.463 & $0.589^{*}$ & $0.737^{*}$ & $0.661^{*}$ & 0.659 & 0.622 & \textbf{0.678} \\
\textbf{ACGAN-Mod-ADC} & $0.526^{*}$ & $0.477^{*}$ & 0.308 & 0.639 & 0.605 & 0.511 & \textbf{0.561} \\
\textbf{ACGAN-Mod-TAC} & $0.416^{*}$ & 0.159 & 0.202 & 0.248 & 0.152 & 0.236 & \textbf{0.431} \\
\textbf{SNGAN} & 0.740  & $0.745^{*}$ & 0.749 & 0.754 & $0.734^{*}$ & 0.744 & \textbf{0.757} \\
\textbf{SAGAN} & 0.781 & 0.745 & $0.784^{*}$ & $0.743^{*}$ & $0.712^{*}$ & 0.753 & \textbf{0.759} \\
\textbf{BigGAN} & $0.733^{*}$ & 0.715 & 0.754 & $0.716^{*}$ & 0.616 & 0.707 & \textbf{0.744} \\
\textbf{BigGAN-Info} & $0.717^{*}$ & $0.774^{*}$ & 0.695 & 0.472 & $0.617^{*}$ & 0.655 & \textbf{0.747} \\
\textbf{BigGAN-ADA} & 0.790 & $0.702^{*}$ & 0.785 & 0.745 & $0.754^{*}$ & 0.755 & \textbf{0.775} \\
\textbf{BigGAN-DiffAug} & 0.648 & $0.769^{*}$ & $0.771^{*}$ & $0.768^{*}$ & $0.802^{*}$ & 0.752 & \textbf{0.784} \\
\textbf{ReACGAN} & 0.758 & 0.728 & $0.738^{*}$ & 0.711 & $0.649^{*}$ & 0.717 & \textbf{0.737} \\
\textbf{ReACGAN-Info} & 0.756 & 0.684 & $0.570^{*}$ & 0.568 & 0.118 & 0.539 & \textbf{0.728} \\
\textbf{ReACGAN-ADA} & 0.764 & 0.754 & $0.804^{*}$ & $0.802^{*}$ & 0.787 & 0.782 & \textbf{0.812} \\
\textbf{ReACGAN-DiffAug} & $0.759^{*}$ & 0.759 & $0.743^{*}$ & 0.764 & 0.775 & 0.760 & \textbf{0.767} \\
\textbf{ReACGAN-ADC} & $0.757^{*}$ & 0.761 & 0.600 & 0.345 & $0.691^{*}$ & 0.631 & \textbf{0.742} \\
\textbf{ReACGAN-TAC} & 0.400 & 0.061 & 0.154 & 0.000 & 0.222 & \textbf{0.167} & 0.000 \\
\textbf{MHGAN} & 0.761 & $0.593^{*}$ & 0.225 & 0.223 & $0.548^{*}$ & 0.470 & \textbf{0.717} \\
\textbf{ContraGAN} & 0.714 & 0.769 & 0.740 & 0.690 & 0.722 & 0.727 & \textbf{0.743} \\
\textbf{StyleGAN2} & 0.764 & 0.625 & $0.716^{*}$ & 0.772 & 0.781 & 0.732 & \textbf{0.748} \\
\textbf{StyleGAN2-Info} & 0.801 & 0.763 & 0.732 & 0.763 & $0.793^{*}$ & 0.770 & \textbf{0.802} \\
\textbf{StyleGAN2-D2D-CE} & 0.808 & \colorbox{yellow}{0.839} &\colorbox{yellow}{0.838} & 0.821 & $0.821^{*}$ & 0.825 & \textbf{0.842} \\
\textbf{StyleGAN2-ADA} & 0.812 & \colorbox{yellow}{$0.835^{*}$} & 0.823 & \colorbox{lime}{0.854} & \colorbox{yellow}{0.843} & 0.833 & \textbf{\underline{0.856}} \\
\textbf{StyleGAN2-DiffAug} & $0.656^{*}$ & 0.719 & 0.765 & 0.680 & 0.479 & 0.660 & \textbf{0.777} \\
\hline
\textbf{Mean (M)} & 0.696 & 0.662 & 0.647 & 0.625 & 0.631 &  &  \\
\textbf{Naive (M)} & \textbf{0.825} & \textbf{0.817} & \textbf{0.807} & \textbf{0.807} & \textbf{0.819} &  &  \\
\bottomrule
\end{tabular}%
\end{adjustbox}
\end{table}

\begin{table}[tb]
\caption{Single GAN g-mean scores for BreastMNIST dataset. We report in bold the experiments where Naive (I) or Naive (M) are above the mean performances of models (Mean (M)) and iterations (Mean (I)), respectively. We underline the best overall experiment.  We highlight in green and yellow the top-1 and top-5 performing GANs, respectively. We denote with * the GANs selected by our multi-objective optimisation.}
\label{tab:single_gan_ultrasound}
\centering
\begin{adjustbox}{width=\columnwidth}
\begin{tabular}{l lllll ll}
\toprule
\textbf{\diagbox[linewidth=0pt]{Models}{Iterations}} & \multicolumn{1}{c}{\textbf{20000}} & \multicolumn{1}{c}{\textbf{40000}} & \multicolumn{1}{c}{\textbf{60000}} & \multicolumn{1}{c}{\textbf{80000}} & \multicolumn{1}{c}{\textbf{100000}} & \textbf{Mean (I)} & \textbf{Naive (I)} \\
\midrule
\textbf{ACGAN-Mod} & $0.472^{*}$ & 0.555 & 0.530 & 0.490 & $0.360^{*}$ & 0.481 & \textbf{0.577} \\
\textbf{ACGAN-Mod-ADC} & 0.588 & 0.408 & $0.532^{*}$ & 0.505 & 0.441 & $0.495^{*}$ & \textbf{0.581} \\
\textbf{ACGAN-Mod-TAC} & 0.407 & 0.000 & 0.423 & 0.397 & 0.000 & 0.245 & \textbf{0.429} \\
\textbf{SNGAN} & 0.053 & 0.445 & $0.408^{*}$ & 0.421 & 0.000 & 0.265 & \textbf{0.554} \\
\textbf{SAGAN} & 0.177 & 0.372 & 0.368 & 0.363 & $0.371^{*}$ & \textbf{0.330} & 0.276 \\
\textbf{BigGAN} & 0.415 & 0.179 & 0.224 & 0.000 & 0.000 & 0.164 & \textbf{0.472} \\
\textbf{BigGAN-Info} & 0.426 & 0.318 & 0.311 & 0.125 & 0.419 & \textbf{0.320} & 0.190 \\
\textbf{BigGAN-ADA} & $0.253^{*}$ & $0.506^{*}$ & 0.512 & 0.549 & 0.456 & 0.456 & \textbf{0.640} \\
\textbf{BigGAN-DiffAug} & 0.508 & 0.514 & $0.448^{*}$ & 0.492 & 0.493 & \textbf{0.491} & 0.462 \\
\textbf{ReACGAN} & 0.436 & $0.079^{*}$ & 0.102 & 0.431 & 0.084 & 0.226 & \textbf{0.397} \\
\textbf{ReACGAN-Info} & 0.468 & $0.194^{*}$ & 0.227 & 0.206 & 0.176 & 0.254 & \textbf{0.513} \\
\textbf{ReACGAN-ADA} & $0.488^{*}$ & 0.518 & $0.483^{*}$ & $0.442^{*}$ & $0.456^{*}$ & 0.477 & \textbf{0.514} \\
\textbf{RaACGAN-DiffAug} & 0.530 & $0.504^{*}$ & 0.427 & $0.352^{*}$ & 0.451 & \textbf{0.453} & 0.373 \\
\textbf{ReACGAN-ADC} & $0.272^{*}$ & $0.336^{*}$ & 0.414 & 0.162 & 0.003 & 0.237 & \textbf{0.336} \\
\textbf{ReACGAN-TAC} & 0.042 & 0.000 & 0.015 & 0.000 & 0.008 & \textbf{0.013} & 0.000 \\
\textbf{MHGAN} & 0.426 & $0.571^{*}$ & $0.441^{*}$ & $0.508^{*}$ & 0.419 & 0.473 & \textbf{0.562} \\
\textbf{ContraGAN} & 0.182 & 0.402 & 0.258 & 0.286 & 0.429 & 0.311 & \textbf{0.432} \\
\textbf{StyleGAN2} & $0.638^{*}$ & 0.665 & 0.546 & $0.640^{*}$ & 0.654 & 0.629 & \textbf{0.695} \\
\textbf{StyleGAN2-Info} & 0.643 & 0.617 & $0.643^{*}$ & 0.624 & 0.644 & 0.634 & \textbf{0.698} \\
\textbf{StyleGAN2-D2D-CE} & 0.648 & $0.674^{*}$ & $0.631^{*}$ & 0.623 & 0.663 & 0.648 & \textbf{0.683} \\
\textbf{StyleGAN2-ADA} & \colorbox{yellow}{$0.684^{*}$} & 0.654 & $0.662^{*}$ & \colorbox{lime}{0.707} & \colorbox{yellow}{0.697} & 0.681 & \textbf{\underline{0.729}} \\
\textbf{StyleGAN2-DiffAug} & 0.655 & $0.676^{*}$ & \colorbox{yellow}{$0.697^{*}$} & $0.676^{*}$ & \colorbox{yellow}{0.702} & 0.681 & \textbf{0.713} \\
\hline
\textbf{Mean (M)} & 0.428 & 0.418 & 0.423 & 0.409 & 0.360 &  &  \\
\textbf{Naive (M)} & \textbf{0.679} & \textbf{0.669} & \textbf{0.630} & \textbf{0.664} & \textbf{0.630} &  &  \\
\bottomrule
\end{tabular}%
\end{adjustbox}
\end{table}

\begin{table}[tb]
\caption{Single GAN g-mean scores for AIforCOVID dataset. We report in bold the experiments where Naive (I) or Naive (M) are above the mean performances of models (Mean (M)) and iterations (Mean (I)), respectively. We underline the best overall experiment.  We highlight in green and yellow the top-1 and top-5 performing GANs, respectively. We denote with * the GANs selected by our multi-objective optimisation.}
\label{tab:single_gan_aiforcovid}
\centering
\begin{adjustbox}{width=\columnwidth}
\begin{tabular}{l lllll ll}
\toprule
\textbf{\diagbox{Models}{Iterations}} & \multicolumn{1}{c}{\textbf{20000}} & \multicolumn{1}{c}{\textbf{40000}} & \multicolumn{1}{c}{\textbf{60000}} & \multicolumn{1}{c}{\textbf{80000}} & \multicolumn{1}{c}{\textbf{100000}} & \textbf{Mean (I)} & \textbf{Naive (I)} \\
\midrule
\textbf{ACGAN-Mod} & $0.000^{*}$ & 0.000 & 0.479 & $0.000^{*}$ & 0.391 & 0.174 & \textbf{0.457} \\
\textbf{ACGAN-Mod-ADC} & $0.418^{*}$ & 0.270 & 0.000 & 0.492 & 0.338 & \textbf{0.304} & 0.275 \\
\textbf{ACGAN-Mod-TAC} & 0.440 & 0.006 & 0.160 & $0.153^{*}$ & 0.044 & 0.160 & \textbf{0.477} \\
\textbf{SNGAN} & 0.374 & 0.409 & 0.355 & $0.424^{*}$ & $0.433^{*}$ & \textbf{0.399} & 0.391 \\
\textbf{SAGAN} & 0.345 & 0.462 & 0.329 & 0.331 & 0.399 & 0.373 & \textbf{0.457} \\
\textbf{BigGAN} & 0.386 & $0.489^{*}$ & $0.386^{*}$ & 0.399 & 0.177 & 0.367 & \textbf{0.502} \\
\textbf{BigGAN-Info} & 0.506 & \colorbox{yellow}{0.533} & $0.506^{*}$ & 0.462 & 0.419 & 0.485 & \textbf{0.546} \\
\textbf{BigGAN-ADA} & 0.341 & $0.478^{*}$ & 0.438 & $0.232^{*}$ & 0.465 & 0.391 & \textbf{0.402} \\
\textbf{BigGAN-DiffAug} & 0.000 & 0.000 & 0.053 & 0.100 & 0.026 & 0.036 & \textbf{0.083} \\
\textbf{ReACGAN} & 0.303 & 0.503 & 0.376 & 0.206 & 0.184 & 0.314 & \textbf{0.470} \\
\textbf{ReACGAN-Info} & 0.209 & $0.239^{*}$ & $0.493^{*}$ & 0.372 & $0.285^{*}$ & 0.319 & \textbf{0.407} \\
\textbf{ReACGAN-ADA} & \colorbox{yellow}{0.569} & 0.408 & 0.264 & 0.442 & 0.464 & 0.429 & \textbf{0.503} \\
\textbf{RaACGAN-DiffAug} & 0.020 & 0.198 & 0.000 & 0.298 & 0.110 & 0.125 & \textbf{0.394} \\
\textbf{ReACGAN-ADC} & 0.406 & 0.374 & 0.472 & 0.327 & 0.247 & 0.365 & \textbf{0.504} \\
\textbf{ReACGAN-TAC} & $0.156^{*}$ & 0.258 & 0.339 & 0.348 & $0.284^{*}$ & 0.277 & \textbf{0.280} \\
\textbf{MHGAN} & $0.456^{*}$ & $0.290^{*}$ & 0.170 & 0.510 & 0.487 & 0.383 & \textbf{0.419} \\
\textbf{ContraGAN} & 0.190 & $0.310^{*}$ & 0.396 & $0.359^{*}$ & $0.388^{*}$ & 0.329 & \textbf{0.440} \\
\textbf{StyleGAN2} & 0.504 & 0.480 & 0.519 & 0.470 & $0.402^{*}$ & 0.475 & \textbf{0.523} \\
\textbf{StyleGAN2-Info} & 0.514 & 0.402 & $0.505^{*}$ & 0.505 & $0.501^{*}$ & 0.486 & \textbf{0.519} \\
\textbf{StyleGAN2-D2D-CE} & 0.349 & $0.489^{*}$ & 0.403 & 0.158 & 0.432 & 0.366 & \textbf{0.422} \\
\textbf{StyleGAN2-ADA} & \colorbox{yellow}{$0.531^{*}$} & {0.452} & $0.515^{*}$ & $0.484^{*}$ & \underline{\colorbox{lime}{$0.588^{*}$}} & 0.514 & \textbf{0.518} \\
\textbf{StyleGAN2-DiffAug} & 0.417 & \colorbox{yellow}{$0.553^{*}$} & 0.310 & 0.341 & 0.321 & 0.388 & \textbf{0.424} \\
\hline
\textbf{Mean (M)} & 0.338 & 0.346 & 0.339 & 0.337 & 0.336 &  &  \\
\textbf{Naive (M)} & \textbf{0.454} & \textbf{0.494} & \textbf{0.408} & \textbf{0.499} & \textbf{0.478} &  &  \\
\bottomrule
\end{tabular}%
\end{adjustbox}
\end{table}

\cref{tab:single_gan_XRay} to~\ref{tab:single_gan_aiforcovid} outline the performance when training the ResNet18 using each GAN separately. 
Each table, displays the 22 GAN models along the rows and the 5 training iterations along the columns, i.e., 20\,000, 40\,000, 60\,000, 80\,000, and 100\,000, resulting in 110 model-iteration pairs.
The results from~\cref{tab:single_gan_XRay,tab:single_gan_ultrasound,tab:single_gan_aiforcovid} include the average performance across iterations for each model (Mean (I)) and the average performance across models for each iteration (Mean (M)). 
The final column and row of the Tables show the performances of Naive (I) and Naive (M) experiments, respectively. 
We highlight in bold the experiment where Naive (I) and Naive (M) outperform Mean (I) and Naive (M), in green and yellow the top-1 and top-5 GANs, respectively and underline the best overall result.
Moreover, we denote with * the set of GANs included in our ensemble solution when using SwAV and for each dataset.

\subsection{Which Model? The Naive (M) experiment}
The Naive (M) experiment moved from the idea that ensembling the synthetic datasets from GANs with a different architecture, adversarial loss, \textit{etc.}, can enhance the diversity and increase downstream model performances.
This approach builds an ensemble using all the 22 GAN models available for each training iteration without any selection criteria. Thus, we have five possible ensembles, one per iteration.
Comparing the last and second-to-last rows (Naive (M) and Mean (M) experiments, respectively) from~\cref{tab:single_gan_XRay,tab:single_gan_ultrasound,tab:single_gan_aiforcovid}, the Naive (M) approach always beats the mean performances of single GAN models, demonstrating the benefits of architectural diversity in GANs ensemble.
It reduces the side effects of synthetic images from GANs for which the training failed, e.g., ReACGAN-TAC in~\cref{tab:single_gan_XRay,tab:single_gan_ultrasound} and BigGAN-DiffAug in~\cref{tab:single_gan_aiforcovid}.

Within this approach, practitioners are alleviated from the problem of verifying whether the adversarial training was successful for each GAN, avoiding the question of "Which GAN should we use?".
Finally, the experiments prove that it is not straightforward for practitioners to define the best iteration to stop adversarial training. 
Indeed we found no trend in the g-mean of Mean (M) and Naive (M) across different iterations.

\subsection{When to Stop Training? The Naive (I) experiment}
The Naive (I) experiment leverages the inner diversity within the different training stages of a GAN to build an ensemble.
This approach builds an ensemble using all the five available GAN snapshots, i.e., 20\,000, 40\,000, 60\,000, 80\,000, 100\,000, in a single adversarial training.
Comparing the last and second-to-last columns (Naive (I) and Mean (I) experiments) from~\cref{tab:single_gan_XRay,tab:single_gan_ultrasound,tab:single_gan_aiforcovid}.
We found that the Naive (I) approach outperforms the average performances of using a single GAN at different iterations with a success rate, i.e. the number of times that Naive (I) beats Mean (I), of 95.45\%,  77.3\% and 90.0\% for PneumoniaMNIST, BreastMNIST, and AIforCOVID, respectively.
Finally, analysing Mean (I) experiments from~\cref{tab:single_gan_XRay,tab:single_gan_ultrasound,tab:single_gan_aiforcovid},
we observe that StyleGAN2-based architectures have higher performances than the others. 

Creating an ensemble using different GAN snapshots during the same training can avoid training diverse GAN models, and it could be beneficial in applications with a limited amount of computation available. 
However, the Naive (I) is less robust to GAN failure than the Naive (M) approach. Indeed, in case of the whole failure of the adversarial training, sampling several iterations does not lead to any benefit in the downstream task, e.g., the performances of ReACGAN-TAC in~\cref{tab:single_gan_XRay,tab:single_gan_ultrasound} are low both for Mean (I) and Naive (I).

\subsection{Comparison with the Pareto optimisation.}

We compared the single GAN experiments, Naive (M) and Naive (I) approaches, against the solution $G^{*}$ selected according to the Oracle and the best backbone (\cref{tab:backbones_analysis}).
Considering the best values obtained from~\cref{tab:single_gan_XRay,tab:single_gan_ultrasound,tab:single_gan_aiforcovid} (underlined values), we point out that neither single GANs, Naive (I) nor Naive (M), can achieve superior performance with respect to the Pareto optimal ensemble.
Indeed, using the ensemble from the Pareto multi-objective optimisation, we achieve a g-mean $0.881$, $0.755$, and $0.612$ compared to $0.856$, $0.729$, and $0.588$ for PneumoniaMNIST, BreastMNIST, and AIforCOVID datasets, respectively.
Thus, combining models and sampling them at different iterations is beneficial for maximising diversity and the ResNet18 performances.

Moreover, to find out the best-performing experiments for each dataset (from~\cref{tab:single_gan_XRay,tab:single_gan_ultrasound,tab:single_gan_aiforcovid}), practitioners need to train 137 ResNet18 (110 experiments for single GAN, five for Naive (M), and 22 for Naive (I)).
Indeed, as shown by the experiments, finding the best-performing GAN ensemble is not straightforward.
In contrast, our approach does not require any posthoc analysis as it searches for the best GANs ensemble with an agnostic procedure, i.e., independent from the downstream task.

\subsection{Insights from single GANs.}

\begin{table}[tb]
\caption{Effect of regularisation ($+$ Info), data augmentation  ($+$ ADA or $+$ DiffAug) and different class conditioning  ($+$ ADC or $+$ TAC) in GAN training.
We count $+1$ if a technique increases the g-mean performance with respect to the baseline, $-1$ otherwise. The possible range for Info, ADA and DiffAug is $[-3,3]$, and $[-2,2]$ for ADC, TAC.}
\label{tab:gan_ranking}
\centering
\begin{adjustbox}{width=\linewidth}
\begin{tabular}{lcccc}
\toprule
 & \textbf{PneumoniaMNIST} & \textbf{BreastMNIST} & \textbf{AIforCOVID} & \textbf{Tot} \\ \midrule
\textbf{+ Info} & $-1$ &$+3$ &$+3$ &$+5$ \\
\midrule
\textbf{+ ADA} &$+3$ &$+3$ &$+3$ & $+9$  \\
\textbf{+ DiffAug} &$+2$ &$+3$ & -3 &$+2$  \\
\midrule
\textbf{+ ADC} &$-2$ &$+2$ &$+2$ &$+2$ \\
\textbf{+ TAC} &$-2$ &$-2$ &$-2$ &$-6$ \\
\bottomrule
\end{tabular}%
\end{adjustbox}
\end{table}

In the search space, we defined GANs with different architectures, conditioning goals, adversarial loss, regularisation, and data-efficient methods.
In~\cref{tab:gan_ranking}, we analyse the effect on GAN training, including the information-theoretic regularisation ($+$ Info), data-efficient training procedures ($+$ ADA or $+$ DiffAug) and different conditioning methods ($+$ ADC or $+$ TAC).
Thus, in~\cref{tab:gan_ranking}, we denote with $+$ when training the GAN with each technique.
For each dataset, we rank a $+1$ if the technique increases the g-mean performances to not using it and $-1$ otherwise.
We compare the GANs that employ the technique against those that do not use it, e.g., BigGAN, StyleGAN2, and ReACGAN against BigGAN-ADA, StyleGAN2-ADA, and ReACGAN-ADA.
Thus, for Info regularisation, ADA and DiffAug, we have a possible range of $[-3, +3]$ and  $[-2, +2]$ for ADC and TAC.
In the last column of~\cref{tab:gan_ranking}, we report the sum across datasets to have a final rank for each method. 

From the first row of~\cref{tab:gan_ranking} emerges that the information-theoretic regularisation is an efficient way to stabilise the GAN training and increase performances. 
Indeed, it achieves a total rank of $+5$ across datasets.
The second and third rows of~\cref{tab:gan_ranking}, show that using ADA as a data-efficient method always improves the baseline with a total rank of $+9$, while DiffAug appears more affected by the type of images used, with a rank of $+2$.
Indeed, it succeeded with PneumoniaMNIST and BreastMNIST datasets while failing with AIforCOVID images.
Turning to different conditioning methods for AC-GAN (last and second-to-last rows in~\cref{tab:gan_ranking}), we found ADC to be a viable choice, with a total rank of $+2$.

\end{document}